\pdfoutput=1

\documentclass[10pt,twocolumn,letterpaper]{article}

\usepackage[table,dvipsnames]{xcolor}  %

\usepackage[pagenumbers]{cvpr}      %

\usepackage[T1]{fontenc}
\usepackage{microtype}
\usepackage{array}
\usepackage{multirow}
\usepackage{makecell}
\usepackage{placeins}
\usepackage{float}       %
\usepackage{tikz}
\usetikzlibrary{arrows.meta,positioning,fit,calc,backgrounds,decorations.pathreplacing,shadings}
\usepackage{adjustbox}

\definecolor{archblue}{RGB}{37,99,235}
\definecolor{archteal}{RGB}{13,148,136}
\definecolor{archorange}{RGB}{234,88,12}
\definecolor{archgold}{RGB}{202,138,4}
\definecolor{archslate}{RGB}{71,85,105}
\definecolor{archviolet}{RGB}{124,58,237}

\newcommand{\events}{\mathcal{E}}
\newcolumntype{L}[1]{>{\raggedright\arraybackslash}p{#1}}

\definecolor{cvprblue}{rgb}{0.21,0.49,0.74}
\usepackage[pagebackref,breaklinks,colorlinks,allcolors=cvprblue]{hyperref}

\def\confName{CVPR}
\def\confYear{2027}

\title{Self-Supervised Continuous-Time Depth Learning and Ego-Motion Estimation from Asynchronous Events}

\author{Sadiq O. Macaulay\\
University of Surrey\\
{\tt\small s.macaulay@surrey.ac.uk}
\and
Dipon Kumar Ghosh\\
University of Surrey\\
{\tt\small d.ghosh@surrey.ac.uk} 
\and
Hadfield Simon\\
University of Surrey\\
{\tt\small s.hadfield@surrey.ac.uk}\\
}

\begin{document}
\makeatletter
\twocolumn[{%
    \@maketitle
    \vspace{6pt}
    \centerline{\resizebox{0.9\textwidth}{!}{    \begin{tikzpicture}[
        font=\Large,
        boxtext/.style={font=\Large\linespread{1}\selectfont},
        topbox/.style={boxtext, rounded corners=6pt, draw=archblue!80!black, fill=archblue!6, very thick, minimum height=3.0cm, align=center, inner sep=5pt, text width=4.0cm},
        midbox/.style={boxtext, rounded corners=6pt, draw=archteal!80!black, fill=archteal!7, very thick, minimum height=3.0cm, align=center, inner sep=5pt, text width=4.0cm},
        geombox/.style={boxtext, rounded corners=6pt, draw=archorange!85!black, fill=archorange!8, very thick, minimum height=3.0cm, align=center, inner sep=5pt, text width=4.0cm},
        outbox/.style={boxtext, rounded corners=6pt, draw=archviolet!85!black, fill=archviolet!8, very thick, minimum height=1.25cm, align=center, inner sep=5pt, text width=4.0cm},
        lossbox/.style={boxtext, rounded corners=6pt, draw=archslate!75!black, fill=archslate!5, very thick, minimum height=2.25cm, align=center, inner sep=5pt, text width=8.0cm},
        arrow/.style={-{Latex[length=3mm,width=2mm]}, very thick, draw=black!75},
        aux/.style={-{Latex[length=2.6mm,width=1.8mm]}, thick, draw=black!55},
        netarrow/.style={-{Latex[length=2.3mm,width=1.6mm]}, thick, draw=black!65},
        stage/.style={text=archslate},
        netcube/.style={rounded corners=4pt, draw=archteal!80!black, fill=archteal!7, very thick, align=center, inner sep=4pt, minimum height=3.0cm, text width=5.4cm, font=\Large\linespread{0.9}\selectfont},
        separator/.style={line width=1.2pt, dash pattern=on 6pt off 4pt},
        group/.style={rounded corners=8pt, draw=black!22, fill=archorange!3, thick, inner sep=5pt},
        eventgroup/.style={rounded corners=8pt, draw=black!22, fill=archblue!3, thick, inner sep=5pt},
        predgroup/.style={rounded corners=8pt, draw=black!22, fill=archblue!2, thick, inner sep=5pt}
    ]

    \node (img1) at (-1.5,6.0) {\includegraphics[scale=0.85]{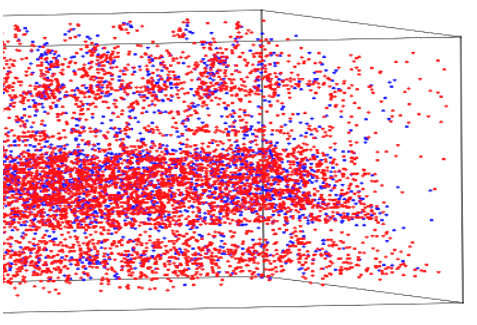}};
    \node (img2) at (15.0,8.0) {\includegraphics[scale=0.72,trim={0cm 0cm 1.5cm 0.0cm},clip]{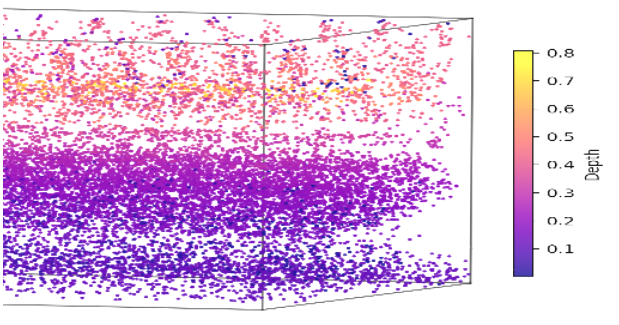}};
    \node (img3) at (14.5,5.0) {\includegraphics[scale=0.72]{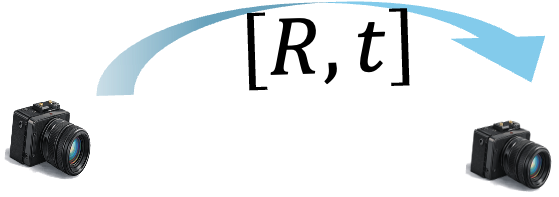}};
    \node (img4) at (24.0,7.0) {\includegraphics[scale=0.8]{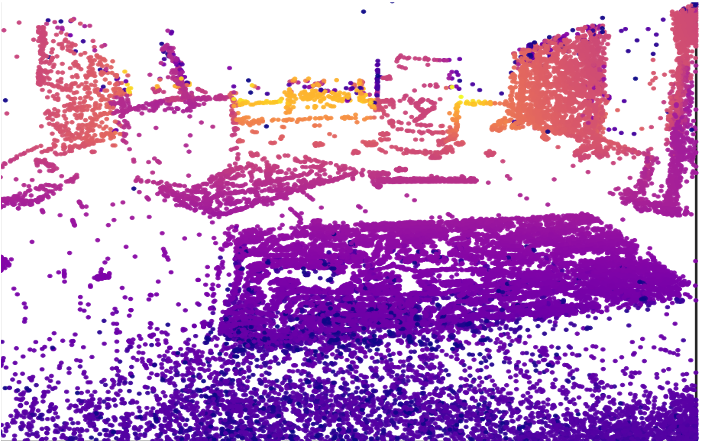}};

    \node[netcube] (net3d) at (6.45,6.45) {\textbf{Event-state depth network}}; %
    \path[draw=archteal!80!black, fill=archteal!14, thick, line join=round]
        (net3d.north west) -- ++(0.38,0.30) -- ($(net3d.north east)+(0.38,0.30)$) -- (net3d.north east) -- cycle;
    \path[draw=archteal!80!black, fill=archteal!20, thick, line join=round]
        (net3d.north east) -- ++(0.38,0.30) -- ($(net3d.south east)+(0.38,0.30)$) -- (net3d.south east) -- cycle;
    \coordinate (netout) at ($(net3d.east)+(0.38,0.15)$);

    \node[stage, font=\large\bfseries] (depthtag) at (15.0,9.75) {Predicted event depth};
    \node[stage, font=\large\bfseries] (posetag) at (15.0,3.75) {Estimated camera motion};

    \draw[
        -{Latex[length=3mm,width=2mm]},
        double,
        double distance=10pt,
        dash pattern=on 3pt off 2pt,
        -{Latex[open,length=7mm,width=10mm]},
        line width=1.5pt
    ] ($(img1.east)+(0.3,0.2)$) -- ($(net3d.west)+(-0.0,-0.2)$);

    \draw[
        -{Latex[length=3mm,width=2mm]},
        double,
        double distance=10pt,
        dash pattern=on 3pt off 2pt,
        -{Latex[open,length=7mm,width=10mm]},
        line width=1.5pt
    ] ($(net3d.east)+(0.4,0.0)$) -- ($(img3.west)+(-1.0,1.5)$);

    \node[topbox] (events) at (-1.3,0.0) {\textbf{Raw Asynchronous event stream}\\ native-time events};

    \node[midbox] (encoder) at (4.0,0.0) {\textbf{Event-state encoder}\\ tokenization + temporal state + local context};

    \node[geombox] (heads) at (9.2,0.0) {\textbf{Depth inference}\\ per-event depth};

    \node[geombox] (render) at (14.4,0.0) {\textbf{Continuous-time warping}\\ interpolate camera motion};

    \node[outbox] (depthout) at (24.0,0.0) {\textbf{Occlusion-aware rendering}\\ target-time depth frame $\hat D_{t^\star}$};

    \node[lossbox] (ssl) at (24.0,-2.8) {\textbf{Self-supervised alignment}\\contrast, time spread, photometric smoothness};

    \node[lossbox] (sup) at (24.0,2.8) {\textbf{Temporally aligned self-supervision}\\ motion-compensated event and rendered-depth losses};

    \begin{scope}[on background layer]
        \node[group, fit=(ssl)(sup)(depthout)] (supervision) {};
        \node[eventgroup, minimum height=6.0cm, fit=(events)(encoder)(heads)(render)] (eventgeo) {};
        \node[predgroup, minimum height=6.0cm, fit=(img2)(img3)(depthtag)(posetag)] (pred) {};

        \draw[separator, draw=archblue!75!black] (1.35,10.35) -- (1.35,-3.55);
        \draw[separator, draw=archorange!85!black] (11.75,10.35) -- (11.75,-3.55);
        \draw[separator, draw=archviolet!80!black] (18.60,10.35) -- (18.60,-3.55);
    \end{scope}

    \node[stage] at ($(supervision.south)+(0,-0.7)$) {\textbf{Rendering and Alignment}};
    \node[stage] at ($(eventgeo.south)+(0,-0.7)$) {\textbf{Native Event Processing}};

    \draw[arrow] (events.east) -- (encoder.west);
    \draw[arrow] (encoder.east) -- (heads.west);
    \draw[arrow] (heads.east) -- (render.west);
    \draw[arrow] (render.east) -- (supervision.west);
    \draw[aux] (depthout.south) -- (ssl.north);
    \draw[aux] (depthout.north) -- (sup.south);
    \draw[arrow] (depthout.east) -| ++(3,2) |- (img4.east);
    \draw[arrow] ($(pred.east)+(0.0,0.25)$) -- (img4.west);

    \end{tikzpicture}
}}
    
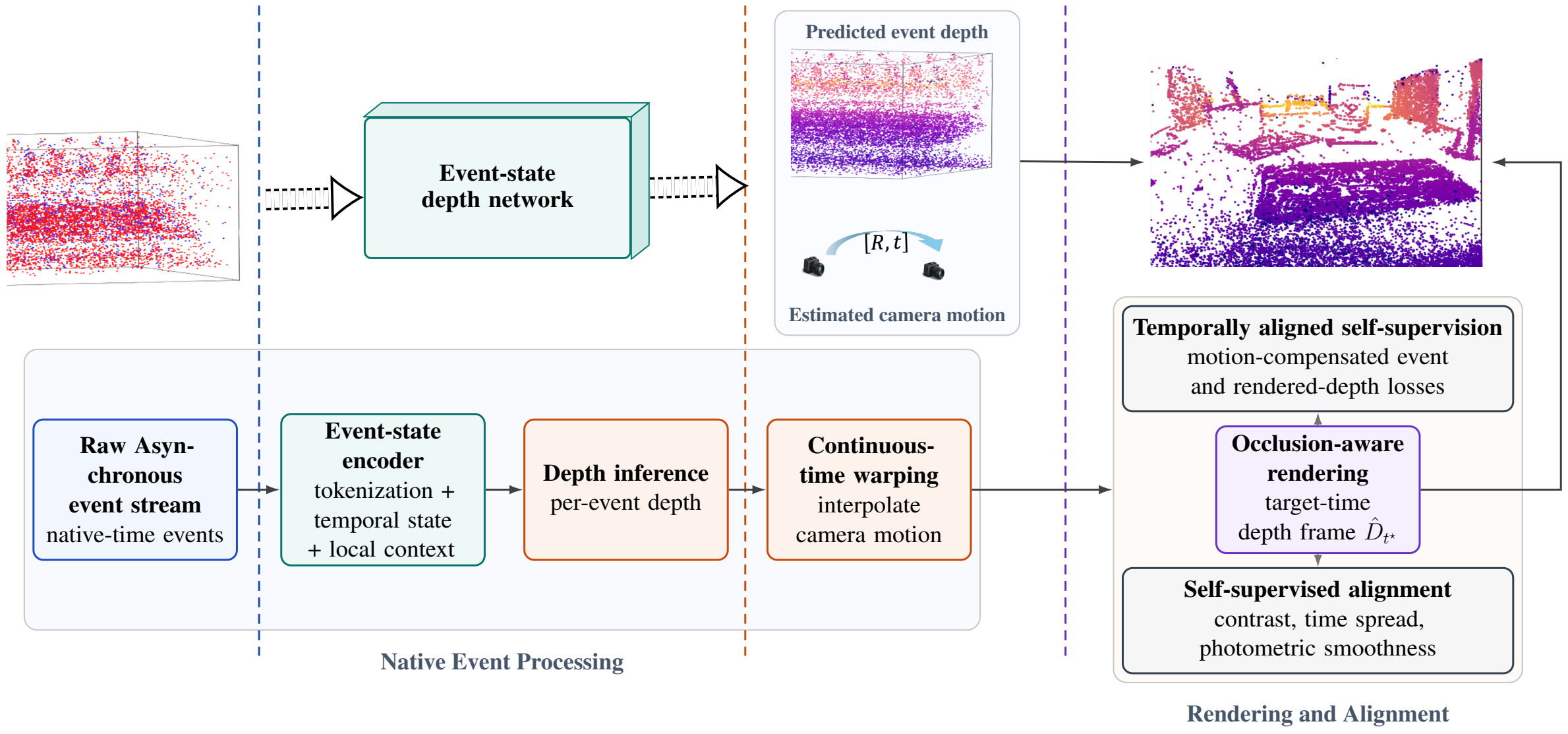
\captionof{figure}{Overview of our method. An event-state network predicts per-event depth. The events are warped in continuous time, their alignment supervises the depth and estimates the camera motion, and an occlusion-aware renderer produces the target-time depth map.}
    \label{fig:intro_pipeline}
    \vspace{6pt}
}]
\makeatother

\begin{abstract}
    Event cameras register brightness changes pixel by pixel with microsecond precision, whereas the dense depth that could supervise them is costly to acquire and exists only at isolated instants. Condensing the stream into frames eases learning, but at the expense of the very timing that sets event sensing apart. We propose a self-supervised approach in which each event is assigned its own depth in continuous time, and the camera motion is recovered from the same observations. Supervision arises from geometry: lifted into 3D by its predicted depth and displaced by the camera motion at the instant it fired, an event should coincide with others produced by the same edge, so any misalignment exposes an error in depth or motion. Differentiable rendering then turns these per-event estimates into dense depth maps at target times within each window, faithful to the events behind them. On MVSEC, the approach outperforms prior event-only self-supervised methods in most sequences and depth ranges, with the widest margins at night, and recovers camera rotation more accurately than an earlier unsupervised event-based method. Ablations isolate the roles of local spatiotemporal context, the temporal backbone, and the alignment objective. These results position motion-compensated event alignment as a principled source of supervision for continuous-time depth and ego-motion estimation.
\end{abstract}

\section{Introduction}
    Each pixel of an event camera fires independently the moment its brightness changes, producing a sparse stream with microsecond temporal resolution and high dynamic range that is well suited to depth estimation in high-speed and low-light regimes \citep{gallego2022event}. In this setting, scene structure and camera motion are coupled prediction problems: each event must be assigned a depth, and the continuous camera motion that carries the same 3D point across the sensor must be known at each event's own timestamp.

    The difficulty is that event streams are irregular in time and sparse in space, whereas dense metric depth is usually available only at image timestamps. Assigning the same endpoint depth map to every event ignores the fact that the camera and the visible geometry evolve within the interval. Most learning pipelines respond to this mismatch by converting events into event frames, voxel grids, or recurrent frame surrogates before estimating geometry \citep{gehrig2019representations,rebecq2019events2video,hidalgo2020e2depth,gehrig2021ramnet,liu2024ereformer}. This suits standard backbones but discards the fine-grained timing that is the main advantage of event sensing.

    We instead derive supervision from the geometry of the measurements themselves. A predicted depth and the camera motion together allow each event to be back-projected at its own timestamp and transported to a common instant. If both are correct, events generated by the same edge arrive at the same place with consistent timestamps. If either is wrong, the events smear across the image. Sharpness after motion compensation is therefore a measure of geometric quality that requires no depth label.

    Our method (Figure~\ref{fig:intro_pipeline}) learns per-event depth from motion-compensated event alignment using a local-global event-state encoder trained by dual-anchor alignment losses. A selective state-space model captures temporal context across the time-ordered events of each window, and a multi-scale local branch captures image-plane neighbourhoods. An occlusion-aware renderer then combines the per-event depths into a target-time map while preserving event-time geometry and gradients to the contributing depths. It anchors each pixel on its most recent observations and combines samples through smooth weights, allowing the rendered map to enter the training objective. A camera-motion head predicts each window's motion, which is also estimated by aligning its events under the predicted depths.

    \paragraph{Contributions.}
        \begin{itemize}
            \item A \emph{self-supervised formulation at event-time granularity}: per-event depth, camera motion evaluated at each event timestamp, and dual-anchor motion-compensated alignment in place of dense depth labels.
            \item A \emph{local-global event-state encoder} that pairs bidirectional Mamba temporal modelling with multi-scale local spatiotemporal aggregation returned to the original event order, a head that predicts a bounded depth per event, and a camera-motion head for the window. Ablations confirm the contributions of the local branch, the state-space backbone, and the alignment objective.
            \item \emph{Geometrically consistent dense depth from per-event depth}: a differentiable occlusion-aware renderer transports each event depth to the target time. A recency-anchored log-depth consensus suppresses older samples that disagree with the most recently observed surface, limiting foreground-background mixing. The target-time map remains consistent with its contributing event depths and passes gradients to each of them. On MVSEC, the rendered depth is the most accurate among event-only self-supervised methods at every 20\,m cut-off and at three of the four 10\,m and 30\,m cut-offs.
        \end{itemize}

\section{Related Work}
    \paragraph{Event representations.} A common route to event-based learning converts the asynchronous stream into a dense surrogate for conventional backbones: learned event representations and voxel-style encodings make the conversion trainable \citep{gehrig2019representations}, event-to-video reconstruction makes events usable in frame-based pipelines \citep{rebecq2019events2video}, and binary event frames suit compact image-like inputs \citep{barchid2022binarep}. The window length, bin count, and normalisation partly determine which motion cues are retained for learning. Event-native alternatives retain more structure: time surfaces encode recent timing \citep{nagata2021timesurface}, voxel graphs preserve sparse neighbourhoods \citep{deng2022voxelgraph}, asynchronous graph networks update features on irregular spatiotemporal graphs \citep{schaefer2022aegnn}, and point-based EventMamba models streams without first forming images \citep{ren2025eventmamba}. Our method follows this event-native direction for per-event depth, keeping events as ordered samples until target-time rendering.

    \paragraph{Geometry from events.} Contrast maximisation provides a unifying view in which correct motion, depth, or flow sharpens an image of warped events \citep{gallego2018contrast}. EVO estimates 6-DoF motion by optimising such alignment \citep{rebecq2017evo}, and stereo and visual-inertial systems add geometric constraints or sensors \citep{zhou2021esvo,rosinol2018ultimate,chen2023esvio,niu2025esvo2}. These methods show the value of continuous-time warping. In our setting, the learned component is the per-event depth field that, with the camera motion, explains the stream.

    \paragraph{Learned and self-supervised event depth.} E2Depth predicts dense monocular depth from accumulated events \citep{hidalgo2020e2depth}, recurrent multimodal networks combine events with frames \citep{gehrig2021ramnet}, and transformer-style recurrent models strengthen temporal aggregation \citep{liu2024ereformer}. Label-free depth learning from video supervises geometry by warping frames and minimising reconstruction error \citep{zhou2017sfmlearner,godard2019monodepth2}, and event warping now drives self-supervised event flow and depth \citep{zhu2018evflownet,zhu2019unsup,zhu2024unsup,hagenaars2025ondeviceselfsup}, in some cases together with ego-motion \citep{zhu2019unsup}. Our contribution is therefore not the warping signal itself but the temporal granularity at which it is applied: existing methods aggregate events into frames or fixed windows and enforce consistency at those discrete anchors, whereas our method applies alignment per event under continuous-time camera motion and predicts depth at event time.

    \paragraph{Sequence models.} Structured state-space models offer linear-time sequence modelling with persistent latent dynamics \citep{gu2022s4}, and selective state-space models such as Mamba add input-dependent updates that scale to long sequences \citep{gu2024mamba}. For event data, this is a middle ground between frame aggregation and global attention: a compact state summarises the time-ordered events of a window, while local spatiotemporal aggregation restores neighbourhood structure around each event.

\section{Method}

    \begin{figure*}[t]
    \centering
    \includegraphics[width=\textwidth]{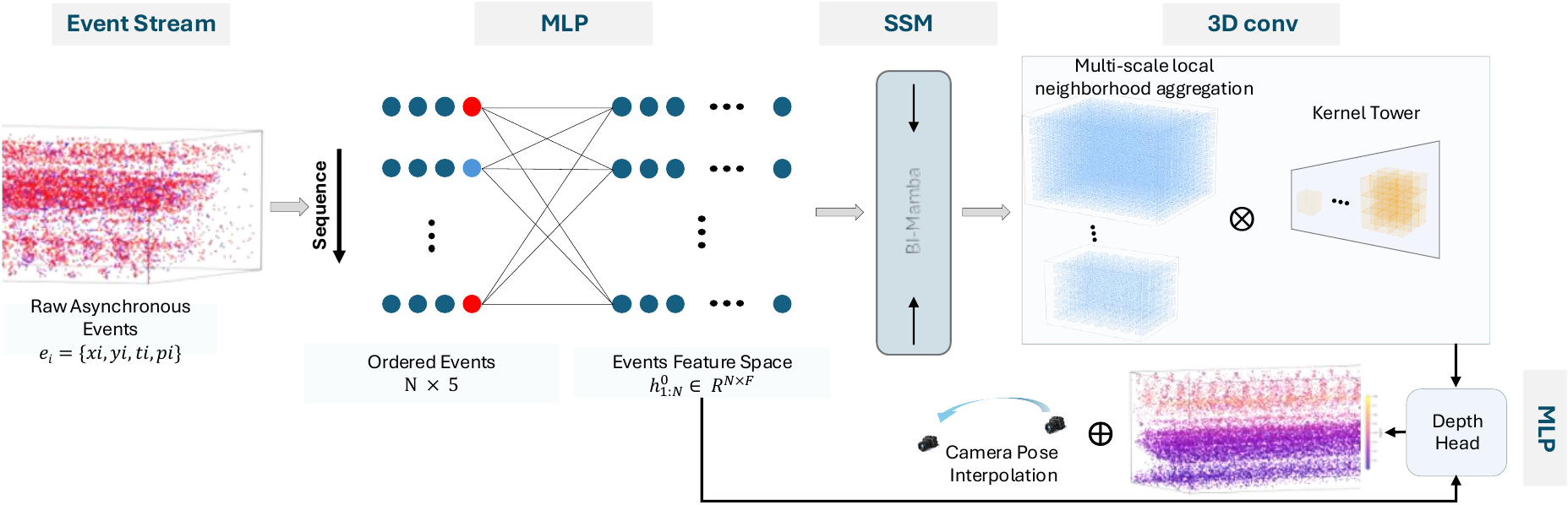}
    \caption{Network architecture: a pointwise event embedding, bidirectional selective state-space modelling, and multi-scale local aggregation feed a skip-assisted per-event depth head.}
    \label{fig:blockflow}
    \end{figure*}

    \subsection{Problem Setting}
    \label{sec:problem}
    A training sample is an event window over $[t_0,t_1]$,
    \begin{equation}
    \events = \{e_i\}_{i=1}^{N}, \qquad
    e_i = (x_i, y_i, \tau_i, p_i, \log \Delta \tau_i),
    \end{equation}
    where $(x_i,y_i)$ is the event location, $\tau_i=(t_i-t_0)/(t_1-t_0)\in[0,1]$ is the normalised event time, $p_i\in\{-1,+1\}$ is polarity, and $\log\Delta\tau_i$ is the logarithm of the elapsed time since the preceding event, which indicates local event density without quantising the stream into temporal bins. The model predicts a positive depth $\hat d_i$ and a log variance $\log\hat\sigma_i^2$ for every event: $\hat d_i$ locates the scene point that triggered $e_i$ along its ray, and $\log\hat\sigma_i^2$ records how strongly that estimate should influence target-time rendering.

    Intrinsics $\mathbf{K}$ come from the calibration \citep{zhu2018mvsec}; metric depth is reserved for evaluation. Endpoint poses $\mathbf{T}_0=\mathbf{I}$ and $\mathbf{T}_1$ follow from the window's camera motion, a constant 6-DoF velocity. Over the short window, translation is interpolated linearly and rotation follows the geodesic between the endpoint orientations, so that $\mathbf{T}_i=(\mathbf{R}_i,\mathbf{t}_i)$ with
    \begin{equation}
    \begin{gathered}
    \mathbf{R}_i=\mathbf{R}_0\exp\!\big(\tau_i\log(\mathbf{R}_0^\top\mathbf{R}_1)\big),\\
    \mathbf{t}_i=(1-\tau_i)\mathbf{t}_0+\tau_i\mathbf{t}_1,
    \end{gathered}
    \label{eq:pose-interp}
    \end{equation}
    written compactly as $\mathbf{T}_i=\mathrm{Interp}_{\mathrm{SE}(3)}(\mathbf{T}_0,\mathbf{T}_1,\tau_i)$. For a target time $t^\star\in[t_0,t_1]$, the method produces two related depth outputs: the set of event-supported depths $\{(x_i,y_i,t_i,\hat d_i,\log\hat\sigma_i^2)\}_{i=1}^{N}$ and a target-time depth map $\hat D_{t^\star}$ rendered from the transported event depths. The formulation assumes calibrated, synchronised sensors, a predominantly static scene, and smooth camera motion over the window.

    \subsection{Local-Global Event-State Encoder}
        Figure~\ref{fig:blockflow} summarises the network: a pointwise event embedding, a bidirectional Mamba stage, one multi-scale local aggregation branch, and a skip-assisted depth head.

        \paragraph{Tokenisation and temporal state.} Coordinates are scaled by the sensor size and each event is mapped to a token by a small MLP, $\mathbf{h}_i^0=\phi_{\mathrm{enc}}(x_i,y_i,\tau_i,p_i,\log\Delta\tau_i)\in\mathbb{R}^{F}$. The operation is pointwise, so each event receives its own feature before any aggregation. The ordered tokens are then processed by bidirectional Mamba blocks \citep{gu2024mamba,ren2025eventmamba},
        \begin{equation}
        \mathbf{h}_{1:N}^{g} = \mathrm{BiMamba}(\mathbf{h}_{1:N}^{0}),
        \end{equation}
        whose forward and reverse states give each event long-range context without the quadratic cost of global attention. Bidirectional processing suits self-supervised training because the complete window is available when the loss is evaluated.

        \paragraph{Hierarchical local aggregation.} Sequence order does not represent image-plane neighbourhoods: consecutive events can originate from distant pixels, and nearby pixels can be separated by many sequence positions. A local branch therefore organises event features into voxelised spatiotemporal towers at several scales, in which factorised temporal and spatial convolutions model change along time and then aggregate nearby image-plane evidence \citep{deng2022voxelgraph,schaefer2022aegnn}. With voxelisation $\mathcal{V}^{(r)}$ and aggregation $\psi^{(r)}$ at scale $r$, the towers $\mathbf{V}^{(r)}=\psi^{(r)}(\mathcal{V}^{(r)}(\events,\mathbf{h}^{g}_{1:N}))$ are returned to the original event order by a skip-attention remapper. Denoting the complete operation by $\Psi_{\mathrm{local}}$,
        \begin{equation}
        \mathbf{h}_{1:N}^{\ell} = \Psi_{\mathrm{local}}(\events, \mathbf{h}_{1:N}^{g}).
        \end{equation}
        Feature-wise linear modulation \citep{perez2017film} conditions the result on the timing descriptor $\mathbf{q}_i=(\tau_i,\log\Delta\tau_i)$, $\widetilde{\mathbf{h}}_i=\boldsymbol{\gamma}(\mathbf{q}_i)\odot[\mathbf{h}_i^g,\mathbf{h}_i^\ell]+\boldsymbol{\beta}(\mathbf{q}_i)$, so that the interpretation of geometric features can adapt to dense event bursts and quiet intervals.

        \paragraph{Depth and uncertainty head.} Token-conditioned gating combines the pointwise token and the contextual feature into a depth logit $\ell_i$ and a log variance, and the depth is bounded to a physical range:
        \begin{equation}
        \begin{gathered}
        (\ell_i,\log\hat\sigma_i^2)=g_d(\mathbf{h}_i^0,\widetilde{\mathbf{h}}_i),\\
        \hat d_i=d_{\min}+(d_{\max}-d_{\min})\operatorname{sigmoid}(\ell_i).
        \end{gathered}
            \label{eq:depth-head}  %
        \end{equation}
        Bounding is valuable early in self-supervised training, when unconstrained depths can send events behind the camera or far outside the image. The uncertainty gives the renderer a learned measure of support. It does not override geometric validity.

        \paragraph{Camera-motion head.} A separate convolutional head predicts the camera motion of a window as a constant 6-DoF velocity. Its input stacks a polarity-separated event voxel grid, the mean inverse predicted depth of the events at each pixel, an event-presence mask, and normalised pixel coordinates, because the image motion caused by ego-motion depends on both depth and image position. Strided convolutions reduce this stack, and the feature map is flattened rather than pooled before the velocity is regressed, so that the spatial pattern of the image motion is retained. The head is trained by event alignment under its own motion with the predicted depths held fixed, together with a penalty for deviating from the velocity that best aligns the window's events.

    \subsection{Continuous-Time Warping and Occlusion-Aware Dense Depth}
    \label{sec:rendering}
        Let $\tilde{\mathbf{x}}_i=(x_i,y_i,1)^\top$ and let $\pi$ denote perspective division. The target pose $\mathbf{T}_\star$ is Equation~\ref{eq:pose-interp} evaluated at $\tau^\star$, and each event is back-projected with its own pose and transported to the target camera,
        \begin{equation}
        \begin{bmatrix}\hat{\mathbf{X}}_i^\star\\1\end{bmatrix}
        =\mathbf{T}_\star^{-1}\mathbf{T}_i
        \begin{bmatrix}\hat d_i\mathbf{K}^{-1}\tilde{\mathbf{x}}_i\\1\end{bmatrix},\quad
        \mathbf{x}_i^\star=\pi(\mathbf{K}\hat{\mathbf{X}}_i^\star),\quad
        \hat z_i^\star=[\hat{\mathbf{X}}_i^\star]_z .
        \label{eq:warp}
        \end{equation}
        Depth enters through the translational part of $\mathbf{T}_\star^{-1}\mathbf{T}_i$: a nearby point moves further on the image than a distant one under the same translation, which is the dependence the self-supervised objective exploits. A sample is valid ($m_i=1$) only when $\hat z_i^\star>0$ and its bilinear support overlaps the image.

        \paragraph{Weighted splatting.} The projected coordinate is generally non-integer, so each event reaches the four nearest pixels $\mathbf{u}$ with bilinear weights $w^{\mathrm{bil}}_{\mathbf{u},i}=\max(0,1-|u_x-x^\star_{i,x}|)\max(0,1-|u_y-x^\star_{i,y}|)$, which are differentiable in the projected coordinate and hence in the predicted depth. Each event also receives a weight
        \begin{equation}
        \begin{gathered}
        w_i=m_i\,\rho_i\,\kappa_i,\qquad \rho_i=e^{-\Delta\tau_i/\tau_r},\\
        \kappa_i=\operatorname{clip}\big(e^{-\log\hat\sigma_i^2},\kappa_{\min},\kappa_{\max}\big),
        \end{gathered}
        \label{eq:render-weight}
        \end{equation}
        where $\Delta\tau_i$ is the normalised time between the event and the endpoint being rendered ($1-\tau_i$ at $t_1$, $\tau_i$ at $t_0$) and $\tau_r$ is a recency time constant. Recency favours the events whose visibility best represents the target view, and the clipped confidence down-weights weakly supported events without letting any single event dominate or vanish. With $w_{\mathbf{u},i}=w^{\mathrm{bil}}_{\mathbf{u},i}w_i$ and $\mathcal{N}(\mathbf{u})$ the events reaching pixel $\mathbf{u}$, the basic renderer returns the weighted mean
        \begin{equation}
        \hat D_{t^\star}(\mathbf{u})=
        \frac{\sum_{i\in\mathcal{N}(\mathbf{u})} w_{\mathbf{u},i}\,\hat z_i^\star}
             {\sum_{i\in\mathcal{N}(\mathbf{u})} w_{\mathbf{u},i}+\epsilon},
        \label{eq:render-mean}
        \end{equation}
        which passes gradients to every contributing event depth.

        \paragraph{Occlusion-aware consensus.} Averaging is appropriate when all samples at a pixel describe one surface, but not at depth discontinuities. Under camera motion, a foreground boundary observed late in the window and a background edge observed earlier can reach the same target pixel, and their mean would not match either. Recency weighting softens this mixing but does not remove it, because older samples of a now-hidden surface still pull the average. The renderer therefore resolves collisions with a recency-anchored consensus in log depth (Figure~\ref{fig:occlusion-rendering}). A short-memory anchor is formed from the most recent samples,
        \begin{equation}
        \ell^{a}(\mathbf{u})=
        \frac{\sum_{i} w^{\mathrm{bil}}_{\mathbf{u},i}\,m_i\rho_i^{a}\kappa_i\log\hat z_i^\star}
             {\sum_{i} w^{\mathrm{bil}}_{\mathbf{u},i}\,m_i\rho_i^{a}\kappa_i+\epsilon},
        \label{eq:render-anchor}
        \end{equation}
        where $\rho_i^{a}$ uses a short time constant $\tau_a$ and the sums run over $\mathcal{N}(\mathbf{u})$. Each sample is compared with the anchor at its own projected location,
        \begin{equation}
        c_i=\exp\!\big(-|\log\hat z_i^\star-\ell^{a}(\mathbf{x}_i^\star)|/\sigma_c\big),
        \label{eq:render-compat}
        \end{equation}
        so that samples consistent with the recently observed surface receive $c_i\approx1$ and samples from another surface $c_i\approx0$. A support estimate $\ell^{s}(\mathbf{u})$ is formed like Equation~\ref{eq:render-anchor} but with a longer time constant $\tau_s$ and each weight multiplied by $c_i$. Its total weight $M(\mathbf{u})$ gives a support confidence $s(\mathbf{u})=1-\exp(-M(\mathbf{u})/M_0)$, and the rendered depth fuses the two estimates,
        \begin{equation}
        \log\hat D_{t^\star}(\mathbf{u})=
        \frac{\alpha\,\ell^{a}(\mathbf{u})+s(\mathbf{u})\,\ell^{s}(\mathbf{u})}{\alpha+s(\mathbf{u})},
        \label{eq:render-fusion}
        \end{equation}
        where $\alpha>0$ sets the influence retained by the anchor. With little compatible history the rendered value follows the recent anchor. Where the same surface has been observed repeatedly, the compatible history refines it, and older samples of a hidden background are gated out rather than averaged into the foreground. The gate $c_i$ is treated as a constant during back-propagation, so gradients act on the depths themselves rather than on their admission to the support. Every step remains a weighted average, and the rendered map stays differentiable with respect to all contributing event depths. Pixels without warped event support remain empty: the renderer does not create measurements on silent surfaces.

\definecolor{occlpurple}{RGB}{82,40,150}
\definecolor{occlblue}{RGB}{25,85,195}
\definecolor{occlcyan}{RGB}{40,180,205}
\definecolor{occlgreen}{RGB}{90,175,90}
\definecolor{occlyellow}{RGB}{242,195,35}
\definecolor{occlorange}{RGB}{238,110,25}
\definecolor{occlred}{RGB}{215,35,25}
\definecolor{occldarkgreen}{RGB}{25,115,55}

\definecolor{plasma0}{HTML}{0D0887}
\definecolor{plasma1}{HTML}{5302A3}
\definecolor{plasma2}{HTML}{8B0AA5}
\definecolor{plasma3}{HTML}{B83289}
\definecolor{plasma4}{HTML}{DB5C68}
\definecolor{plasma5}{HTML}{F48849}
\definecolor{plasma6}{HTML}{FEBD2A}
\definecolor{plasma7}{HTML}{F0F921}

\newcommand{\plasmacolor}[1]{%
    \pgfmathsetmacro{\pv}{max(0,min(1,#1))}%
    \pgfmathtruncatemacro{\pi}{min(6,floor(7*\pv))}%
    \pgfmathsetmacro{\pmix}{100*(7*\pv-\pi)}%
    \pgfmathsetmacro{\pweight}{100-\pmix}%
    \ifcase\pi
        \colorlet{eventcolor}{plasma0!\pweight!plasma1}%
    \or
        \colorlet{eventcolor}{plasma1!\pweight!plasma2}%
    \or
        \colorlet{eventcolor}{plasma2!\pweight!plasma3}%
    \or
        \colorlet{eventcolor}{plasma3!\pweight!plasma4}%
    \or
        \colorlet{eventcolor}{plasma4!\pweight!plasma5}%
    \or
        \colorlet{eventcolor}{plasma5!\pweight!plasma6}%
    \or
        \colorlet{eventcolor}{plasma6!\pweight!plasma7}%
    \fi
}

\pgfdeclareverticalshading{plasmashading}{100bp}{%
    color(0bp)=(plasma0);
    color(14.3bp)=(plasma1);
    color(28.6bp)=(plasma2);
    color(42.9bp)=(plasma3);
    color(57.1bp)=(plasma4);
    color(71.4bp)=(plasma5);
    color(85.7bp)=(plasma6);
    color(100bp)=(plasma7)
}

\begin{figure*}[t]
    \centering

    \begin{adjustbox}{max width=\textwidth}
    \begin{tikzpicture}[
        x=1cm,
        y=1cm,
        font=\sffamily,
        >=Latex,
        stage/.style={
            font=\scriptsize\bfseries,
            text=occlpurple,
            align=center
        },
        number/.style={
            circle,
            fill=occlpurple,
            text=white,
            font=\scriptsize\bfseries,
            inner sep=2.2pt
        },
        rednumber/.style={
            circle,
            fill=occlred,
            text=white,
            font=\scriptsize\bfseries,
            inner sep=2.2pt
        },
        greennumber/.style={
            circle,
            fill=occldarkgreen,
            text=white,
            font=\scriptsize\bfseries,
            inner sep=2.2pt
        },
        majorarrow/.style={
            -{Latex[length=3mm,width=2.4mm]},
            draw=occldarkgreen,
            line width=1.5pt
        },
        warparrow/.style={
            dashed,
            line width=0.65pt,
            -{Latex[length=1.6mm,width=1.1mm]}
        },
        panel/.style={
            rounded corners=3pt,
            line width=0.55pt,
            inner sep=4pt
        }
    ]

\node[number] at (0.25,6.85) {1};

\node[
    stage,
    anchor=west
] at (0.55,6.85)
{\shortstack{\scriptsize Back-projected events\\
\tiny(native timestamps)}};

\pgfmathsetseed{20}

\def\xmin{0.50}
\def\xmax{2.95}
\def\ymin{2.60}
\def\ymax{5.85}

\def\rmin{0.035}
\def\rmax{0.120}

\def\nevents{70}

\foreach \i in {1,...,\nevents} {%

    \pgfmathsetmacro{\px}{
        \xmin + (\xmax-\xmin)*rnd
    }
    \pgfmathsetmacro{\py}{
        \ymin + (\ymax-\ymin)*rnd
    }

    \pgfmathsetmacro{\pr}{
        \rmin + (\rmax-\rmin)*rnd
    }

    \pgfmathsetmacro{\pvalue}{
        (\i-1)/(\nevents-1)
    }

    \plasmacolor{\pvalue}

    \fill[eventcolor]
        (\px,\py) circle (\pr);
}

\def\cbxmin{0.10}
\def\cbxmax{0.26}

\shade[
    shading=plasmashading
]
(\cbxmin,\ymin)
rectangle
(\cbxmax,\ymax);

\draw[
    black,
    line width=0.35pt
]
(\cbxmin,\ymin)
rectangle
(\cbxmax,\ymax);

\foreach \v/\lab in {
    0.00/{near},
    1.00/{far}
}{
    \pgfmathsetmacro{\cby}{
        \ymin + \v*(\ymax-\ymin)
    }

    \draw[
        black,
        line width=0.35pt
    ]
    (\cbxmin,\cby) -- ++(-0.07,0);

    \node[
        anchor=east,
        font=\scriptsize
    ]
    at ({\cbxmin-0.10},\cby)
    {\lab};
}

\node[
    rotate=90,
    font=\scriptsize,
    anchor=south
]
at (
    {\cbxmin},
    {(\ymin+\ymax)/2}
)
{depth};

\draw[
    dashed,
    draw=black!25,
    line width=0.55pt
]
(3.72,1.75) -- (3.72,7.05);

\node[number] at (4.2,6.85) {2};

\node[
    stage,
    font=\scriptsize\bfseries,
    anchor=west
] at (4.4,6.85)
{\shortstack{Warp to target time\\
and project to pixel $\mathbf{u}$}};

\coordinate (planeBL) at (6.25-0.70,3.35);
\coordinate (planeBR) at (7.40-0.70,3.10);
\coordinate (planeTR) at (7.40-0.70,5.45);
\coordinate (planeTL) at (6.25-0.70,5.80);

\path[
    draw=occlorange,
    fill=occlorange!6,
    line width=0.9pt
]
(planeBL) -- (planeBR) -- (planeTR) -- (planeTL) -- cycle;

\foreach \f in {0.1,0.2,0.3,0.4,0.5,0.6,0.7,0.8,0.9}{
    \draw[
        occlorange!20,
        line width=0.35pt
    ]
    ($(planeBL)!\f!(planeTL)$)
    --
    ($(planeBR)!\f!(planeTR)$);
}

\foreach \f in {0.25,0.5,0.75}{
    \draw[
        occlorange!20,
        line width=0.35pt
    ]
    ($(planeBL)!\f!(planeBR)$)
    --
    ($(planeTL)!\f!(planeTR)$);
}

\node[
    draw=black,
    fill=white,
    minimum width=0.19cm,
    minimum height=0.19cm,
    inner sep=0pt,
    line width=0.75pt
] (pixelu) at (6.96-0.70,4.31) {};

\node[
    font=\small\bfseries,
    above=1mm of pixelu
] {$\mathbf{u}$};

\plasmacolor{0.92}\colorlet{selA}{eventcolor}
\plasmacolor{0.78}\colorlet{selB}{eventcolor}
\plasmacolor{0.64}\colorlet{selC}{eventcolor}
\plasmacolor{0.48}\colorlet{selD}{eventcolor}
\plasmacolor{0.30}\colorlet{selE}{eventcolor}
\plasmacolor{0.12}\colorlet{selF}{eventcolor}

\coordinate (srcA) at (2.80,4.30);
\coordinate (srcB) at (2.92,4.18);
\coordinate (srcC) at (2.70,4.15);
\coordinate (srcD) at (2.86,4.02);
\coordinate (srcE) at (2.62,3.96);
\coordinate (srcF) at (2.76,3.84);

\coordinate (warpA) at (4.12,4.95);
\coordinate (warpB) at (4.18,4.68);
\coordinate (warpC) at (4.26,4.42);
\coordinate (warpD) at (4.34,4.16);
\coordinate (warpE) at (4.42,3.92);
\coordinate (warpF) at (4.50,3.70);

\def\rA{0.045}
\def\rB{0.055}
\def\rC{0.065}
\def\rD{0.075}
\def\rE{0.085}
\def\rF{0.095}

\tikzset{
    smoothwarp/.style={
        warparrow,
        line cap=round,
        line join=round
    }
}

\fill[selA] (srcA)  circle (\rA);
\fill[selB] (srcB)  circle (\rB);
\fill[selC] (srcC)  circle (\rC);
\fill[selD] (srcD)  circle (\rD);
\fill[selE] (srcE)  circle (\rE);
\fill[selF] (srcF)  circle (\rF);

\draw[smoothwarp, draw=selA]
(srcA)
.. controls (3.15,4.55) and (3.72,4.90) ..
(warpA)
.. controls (4.70,5.02) and (5.72-0.70,4.88) ..
(pixelu.west);

\draw[smoothwarp, draw=selB]
(srcB)
.. controls (3.18,4.38) and (3.78,4.66) ..
(warpB)
.. controls (4.76,4.78) and (5.75-0.70,4.62) ..
(pixelu.west);

\draw[smoothwarp, draw=selC]
(srcC)
.. controls (3.16,4.24) and (3.86,4.40) ..
(warpC)
.. controls (4.82,4.48) and (5.80-0.70,4.42) ..
(pixelu.west);

\draw[smoothwarp, draw=selD]
(srcD)
.. controls (3.24,4.00) and (3.94,4.10) ..
(warpD)
.. controls (4.88,4.18) and (5.84-0.70,4.26) ..
(pixelu.west);

\draw[smoothwarp, draw=selE]
(srcE)
.. controls (3.12,3.84) and (4.02,3.88) ..
(warpE)
.. controls (4.94,3.96) and (5.90-0.70,4.10) ..
(pixelu.west);

\draw[smoothwarp, draw=selF]
(srcF)
.. controls (3.18,3.72) and (4.10,3.66) ..
(warpF)
.. controls (5.00,3.74) and (5.96-0.70,3.96) ..
(pixelu.west);

\fill[selA] (warpA) circle (\rA);
\fill[selB] (warpB) circle (\rB);
\fill[selC] (warpC) circle (\rC);
\fill[selD] (warpD) circle (\rD);
\fill[selE] (warpE) circle (\rE);
\fill[selF] (warpF) circle (\rF);

\draw[
    dashed,
    draw=black!50
]
(pixelu.south) -- ++(0,-2.08);

\node[
    font=\scriptsize
] at (6.96-0.70,2.05)
{$t_n$};

\draw[
    dashed,
    draw=occlorange,
    line width=0.55pt
]
(8.15,1.75) -- (8.15,7.05);

\draw[majorarrow]
(8.55,4.35-0.2) -- (7.50,4.35-0.2);

\begin{scope}[xshift=-1.55cm]

\node[rednumber] at (12.0,6.85) {3A};
\node[font=\scriptsize\bfseries,anchor=west,text=black!75] at (12.2,6.85) {Recent anchor};

\draw[-{Latex[length=1.7mm]},line width=0.55pt] (11.85,5.02) -- (14.60,5.02);
\draw[-{Latex[length=1.7mm]},line width=0.55pt] (11.85,5.02) -- (11.85,6.45);

\node[font=\scriptsize,below] at (13.20,5.00) {log depth};
\node[font=\scriptsize,rotate=90] at (11.56,5.77) {weight};
\node[font=\scriptsize,below] at (12.02,4.98) {near};
\node[font=\scriptsize,below] at (14.38,4.98) {far};

\foreach \x/\h/\c in {
    12.10/0.98/selF,
    12.30/0.80/selE,
    12.50/0.60/selD,
    13.72/0.12/selC,
    13.96/0.08/selB,
    14.20/0.05/selA%
}{
    \draw[\c,line width=1.0pt] (\x,5.02) -- (\x,{5.02+\h});
    \fill[white,draw=\c,line width=0.75pt] (\x,{5.02+\h}) circle (0.045);
}

\draw[dashed,draw=occlred!80!black,line width=0.65pt] (12.27,5.02) -- (12.27,6.22);
\node[text=occlred!80!black,font=\scriptsize\bfseries,anchor=west] at (12.62,6.08) {anchor};

\node[greennumber] at (12.0,3.9) {3B};
\node[font=\scriptsize\bfseries,anchor=west,text=black!75] at (12.2,3.9) {Gate and fuse};

\draw[-{Latex[length=1.7mm]},line width=0.55pt] (11.85,2.05) -- (14.60,2.05);
\draw[-{Latex[length=1.7mm]},line width=0.55pt] (11.85,2.05) -- (11.85,3.40);

\node[font=\scriptsize,below] at (13.20,2.03) {log depth};
\node[font=\scriptsize,rotate=90] at (11.48,2.73) {\shortstack{support\\weight}};
\node[font=\scriptsize,below] at (12.03,2.01) {near};
\node[font=\scriptsize,below] at (14.37,2.01) {far};

\fill[occldarkgreen!10,domain=11.87:14.50,samples=90]
    (11.87,2.05) -- plot (\x,{2.05+1.10*exp(-abs(\x-12.27)/0.32)}) -- (14.50,2.05) -- cycle;
\draw[occldarkgreen!80!black,line width=0.65pt,domain=11.87:14.50,samples=90]
    plot (\x,{2.05+1.10*exp(-abs(\x-12.27)/0.32)});

\foreach \x/\h/\c in {
    12.10/0.82/selF,
    12.30/0.88/selE,
    12.50/0.62/selD%
}{
    \draw[\c,line width=1.0pt] (\x,2.05) -- (\x,{2.05+\h});
    \fill[white,draw=\c,line width=0.75pt] (\x,{2.05+\h}) circle (0.045);
}
\foreach \x/\c in {13.72/selC, 13.96/selB, 14.20/selA}{
    \draw[\c!45,line width=0.8pt] (\x,2.05) -- (\x,2.15);
    \node[text=black!55,font=\tiny\bfseries] at (\x,2.32) {$\times$};
}
\node[text=black!60,font=\tiny,align=center] at (13.96,2.78) {\shortstack{hidden surface\\suppressed}};

\draw[dashed,draw=occldarkgreen!80!black,line width=0.65pt] (12.29,2.05) -- (12.29,3.30);
\node[text=occldarkgreen!80!black,font=\scriptsize\bfseries,anchor=west] at (12.62,3.22) {$\hat D(\mathbf{u})$};

\begin{scope}[on background layer]

\node[
    panel,
    draw=occlred!35,
    fill=occlred!1,
    fit={(11.0,4.72) (15.00,6.35)}
] {};

\node[
    panel,
    draw=occldarkgreen!40,
    fill=occldarkgreen!1,
    fit={(11.0,1.70) (15.00,3.45)}
] {};

\end{scope}

\end{scope}

\draw[
    densely dashed,
    draw=occlred!85!black,
    line width=0.8pt,
    -{Latex[length=2.0mm,width=1.5mm]}
]
(9.30,1.62) -- (2.05,1.62) -- (2.05,2.40);

\node[
    font=\scriptsize,
    text=occlred!85!black,
    anchor=north
] at (4.60,1.58)
{gradient reaches every event depth $\hat d_i$};

\begin{scope}[shift={(0,-0.25)}]

\begin{scope}[on background layer]
    \node[
        draw=black!20,
        fill=black!1,
        rounded corners=3pt,
        minimum width=10.2cm,
        minimum height=1.45cm
    ] at (6.70,0.7) {};
\end{scope}

\node[
    font=\scriptsize\bfseries
] at (3.10,1.25)
{Recency};

\fill[gray!45] (2.25,0.90) circle (0.035);
\fill[gray!45] (2.65,0.90) circle (0.060);
\fill[gray!45] (3.10,0.90) circle (0.090);
\fill[gray!45] (3.60,0.90) circle (0.125);

\draw[
    -{Latex[length=1.5mm]},
    line width=0.45pt
]
(2.15,0.55) -- (3.85,0.55);

\node[
    font=\scriptsize,
    below=0.5mm
] at (2.20,0.55)
{older};

\node[
    font=\scriptsize,
    below=0.5mm
] at (3.75,0.55)
{recent};

\draw[black!15] (4.30,0.0) -- (4.30,1.45);

\node[
    font=\scriptsize\bfseries
] at (5.25,1.25)
{Target pixel};

\draw[
    rounded corners=1pt,
    draw=black!65
]
(5.02,0.68) rectangle (5.47,1.03);

\draw[black!15] (6.20,0.0) -- (6.20,1.45);

\node[
    font=\scriptsize\bfseries,
    text=occldarkgreen,
    anchor=center,
    align=center,
] at (8.95,1.27)
{Occlusion-aware rendering};

\node[
    font=\scriptsize,
    align=center,
    anchor=center
] at (8.95,0.62)
{\shortstack{recent events anchor the visible surface.\\
Compatible history refines it, hidden-surface\\
samples are suppressed}};

\end{scope}

\node[
    font=\tiny,
    text=black!55,
    align=right
] at (5.5,2.8)
{\parbox[t][\height][t]{3cm}{\raggedright Projection uses\\bilinear splatting where\\multiple projected event depths\\fall on the same pixel}};

\node[
    font=\tiny,
    text=black!55,
    align=right
] at (8.5,3.55)
{\parbox[t][\height][t]{3cm}{\raggedright The pixel value is a\\weighted average of the\\participating event depths.}};

    \end{tikzpicture}
    \end{adjustbox}

    \caption{
        Occlusion-aware dense depth from per-event depth. (1)~Events are back-projected with their predicted depths and
        (2)~transported and splatted to pixel~$\mathbf{u}$ at the target time. (3A)~The most recent samples form a log-depth
        anchor for the visible surface. (3B)~Older samples are admitted by their agreement with the anchor, so hidden
        surfaces are suppressed. Every step is a weighted average, so gradients reach each event depth.
    }
    \label{fig:occlusion-rendering}

\end{figure*}
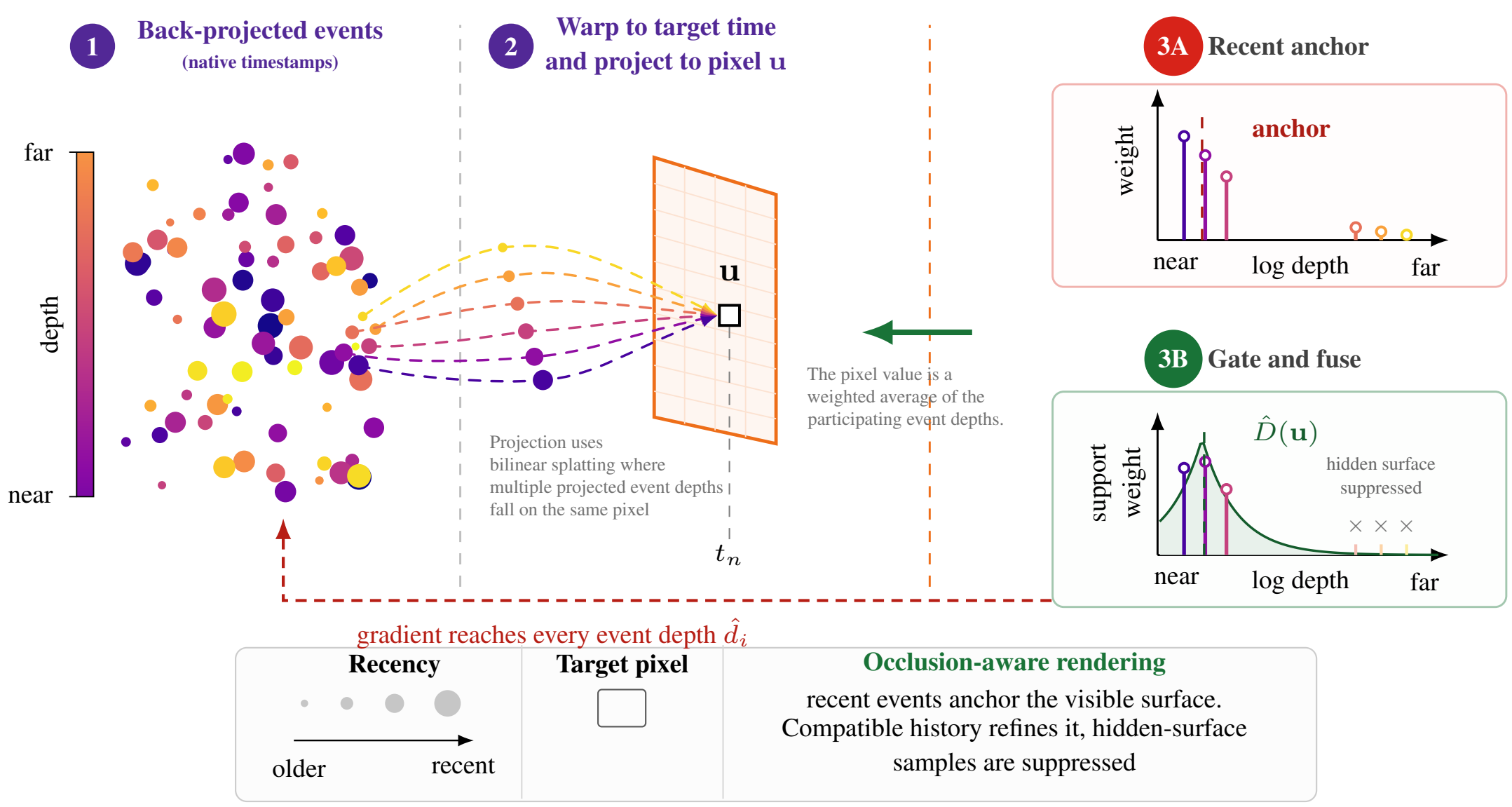

    \subsection{Self-Supervised Objectives}
    \label{sec:losses}
        A depth hypothesis is evaluated by how sharply and consistently it aligns the asynchronous events under the camera motion of the window \citep{gallego2018contrast,zhu2018evflownet,zhu2019unsup}. The model is trained with
        \begin{equation}
            \mathcal{L}_{\mathrm{ssl}} =
            \lambda_{\mathrm{sharp}}\mathcal{L}_{\mathrm{sharp}}
            + \lambda_{\mathrm{time}}\mathcal{L}_{\mathrm{time}}
            + \lambda_{\mathrm{oob}}\mathcal{L}_{\mathrm{oob}}
            + \lambda_{\mathrm{smooth}}\mathcal{L}_{\mathrm{smooth}},
            \label{eq:total-loss}
        \end{equation}
        evaluated after back-projection, pose transport, and splatting, so gradients travel from target-time alignment back to every event-depth logit. That motion is the best-aligning velocity $\bar{\boldsymbol{\xi}}$ under the fixed depths, and a weak gauge term pins the unobservable scale. The alignment terms use the dual anchors $\mathcal{A}=\{t_0,t_1\}$. Let $\mathbf{u}_i^{a}$ be the coordinate of $e_i$ warped to anchor $a$ and $b^{a}_{\mathbf{u},i}$ its bilinear contribution to pixel $\mathbf{u}$. The signed image of warped events and the count image are
        \begin{equation}
            I_a(\mathbf{u})=\sum_{i=1}^{N} p_i\, b^{a}_{\mathbf{u},i}, \qquad
            C_a(\mathbf{u})=\sum_{i=1}^{N} b^{a}_{\mathbf{u},i}.
            \label{eq:iwe}  %
        \end{equation}

        \paragraph{Event sharpness.} Correct depth concentrates events from one physical edge into a few pixels, so the contrast of the warped-event image is maximised:
        \begin{equation}
            \mathcal{L}_{\mathrm{sharp}}=-\sum_{a\in\mathcal{A}}\mathrm{Var}_{\mathbf{u}}\big(I_a(\mathbf{u})\big).
            \label{eq:sharp}  %
        \end{equation}

        \paragraph{Temporal dispersion.} Spatial concentration alone admits degenerate warps that compress events from different times and surfaces into one region. The temporal term penalises the count-weighted within-pixel variance of warped timestamps,
        \begin{equation}
        \begin{gathered}
            \mathcal{L}_{\mathrm{time}}=
            \frac{\sum_{a}\sum_{\mathbf{u}} C_a(\mathbf{u})\,\sigma_a^2(\mathbf{u})}
                 {\sum_{a}\sum_{\mathbf{u}} C_a(\mathbf{u})+\epsilon},\\
            \sigma_a^2(\mathbf{u})=\frac{\sum_i \tau_i^2\, b^{a}_{\mathbf{u},i}}{C_a(\mathbf{u})+\epsilon}-\bar\tau_a^2(\mathbf{u}),
        \end{gathered}
            \label{eq:time}  %
        \end{equation}
        with mean warped time $\bar\tau_a(\mathbf{u})=\sum_i \tau_i\, b^{a}_{\mathbf{u},i}/(C_a(\mathbf{u})+\epsilon)$. Weighting by $C_a$ keeps pixels with negligible support from dominating. Requiring both sharpness and low temporal spread reduces the set of degenerate warps that can satisfy the objective.

        \paragraph{Out-of-bounds and smoothness.} The out-of-bounds term penalises warps that move events outside the sensor rectangle of size $\mathbf{s}=(W,H)^\top$. With the per-event excursion $\boldsymbol{\delta}_i^{a}=\mathrm{ReLU}(-\mathbf{u}_i^{a})+\mathrm{ReLU}(\mathbf{u}_i^{a}-\mathbf{s})$, evaluated before invalid samples are discarded,
        \begin{equation}
            \mathcal{L}_{\mathrm{oob}}=\frac{1}{N|\mathcal{A}|}\sum_{a\in\mathcal{A}}\sum_{i=1}^{N}\big\|\boldsymbol{\delta}_i^{a}\big\|_2^2 .
            \label{eq:oob}  %
        \end{equation}
        The depth rendered at the window end, $\hat D_{t_1}$, is regularised by an edge-aware smoothness term whose edges come from the events. Let $E$ be the image of the in-view events warped to $t_1$ (one-pixel Gaussian blur, divided by its 99th percentile) and $q=\hat D_{t_1}^{-1}/\overline{\hat D_{t_1}^{-1}}$ the rendered inverse depth divided by its mean. Over the set $\mathcal{P}$ of adjacent pixel pairs that are both rendered,
        \begin{equation}
            \mathcal{L}_{\mathrm{smooth}}=\frac{1}{|\mathcal{P}|}\sum_{(\mathbf{u},c)\in\mathcal{P}}
            \big|\partial_c q(\mathbf{u})\big|\,
            e^{-|\partial_c E(\mathbf{u})|},
            \label{eq:smooth}  %
        \end{equation}
        which stabilises regions with limited event support while allowing discontinuities at moving edges. The mean normalisation and the average over pairs make the term independent of the global scale and of the number of rendered pixels. Because the rendered map is a weighted average of transported event depths, this term also passes gradients to the per-event depths; $E$ receives none.

\section{Experiments}
\label{sec:experiments}

    The evaluation tests four claims. First, event alignment can supervise a useful continuous-time depth field without dense depth targets. Second, occlusion-aware rendering converts that field into target-time maps that are competitive under conventional depth metrics. Third, the local-global encoder is both effective and practical for long event windows. Fourth, the learned per-event depth is accurate enough for the camera motion to be recovered from the same events. Metric depth maps enter only after inference, when the rendered predictions are compared with the benchmark reference.

    \subsection{Setup}
        \paragraph{Dataset and windows.} We use MVSEC, which provides synchronised events, image frames, metric depth, camera poses, and calibration \citep{zhu2018mvsec}. The model is trained on \texttt{outdoor\_day2}. Quantitative evaluation uses \texttt{outdoor\_day1} and the three outdoor night sequences, and the qualitative examples come from \texttt{outdoor\_day1} and \texttt{outdoor\_night1}. Because the formulation assumes a predominantly static scene (Section~\ref{sec:problem}), in which the camera motion alone explains how events move across the image, segments of \texttt{outdoor\_day2} with considerable independent object motion are cropped out of the training data. The test sequences are evaluated on every frame, moving objects included. Each sample spans 50\,ms and contains at most $30{,}000$ events. This interval accumulates local spatiotemporal structure while keeping endpoint-pose interpolation a reasonable approximation to the trajectory. Windows exceeding the budget are subsampled in a time-stratified manner, which maintains coverage of the complete interval. Shorter windows are padded with an event mask, so padding cannot contribute to the state, the splatting weights, or the losses. Endpoint metric depth maps are sampled only for evaluation and are not used in Equation~\ref{eq:total-loss}.

        \paragraph{Training.} The model is trained end to end with Equation~\ref{eq:total-loss} using Adam at an initial learning rate of $10^{-5}$ under a ReduceLROnPlateau schedule, with batch size 4 for 50 epochs on an NVIDIA GeForce RTX 2080 Ti. The loss weights are $\lambda_{\mathrm{sharp}}=1.0$, $\lambda_{\mathrm{time}}=0.5$, $\lambda_{\mathrm{oob}}=0.1$, and $\lambda_{\mathrm{smooth}}=0.2$. The recency time constant is $\tau_r=0.2$, and depth is bounded to $[d_{\min},d_{\max}]=[0.1,80.0]$\,m.

        \paragraph{Baselines.} The comparison includes three event-only self-supervised references: \citet{zhu2019unsup} jointly learn optical flow, depth, and ego-motion from events. EMoDepth (E) \citep{zhu2024unsup} is the configuration of EMoDepth's supervision study whose self-supervisory signal comes from event voxels alone. \citet{hagenaars2025ondeviceselfsup} study low-latency on-device monocular depth from events. EMoDepth (E+I) is the main EMoDepth model: it also infers depth from events alone, but is trained with a cross-modal consistency loss on aligned intensity frames. It is retained as a cross-modal reference and excluded from the event-only ranking, so that gains attributable to image supervision are not confused with gains in asynchronous event modelling. All methods are compared at the depth-map level: our per-event depth predictions are transported using the motion predicted by the camera-motion head and rendered at the reference timestamp.

        \paragraph{Metric and scale alignment.} Self-supervised monocular estimates retain a global scale ambiguity, so, following standard practice \citep{zhou2017sfmlearner,godard2019monodepth2}, each rendered prediction is aligned to the ground truth by one median scale factor,
        \begin{equation}
            s=\frac{\mathrm{median}\{\,D_{\mathrm{gt}}(\mathbf{u}): \mathbf{u}\in\Omega_v\,\}}
                   {\mathrm{median}\{\,\hat D_{t^\star}(\mathbf{u}): \mathbf{u}\in\Omega_v\,\}},
            \label{eq:median-scale}  %
        \end{equation}
        where $\Omega_v$ contains the pixels above the vehicle hood with valid ground-truth depth up to 30\,m that receive a transported event. The same factor is used at every cut-off, and the mean absolute error between $s\hat D_{t^\star}$ and $D_{\mathrm{gt}}$ is reported over the pixels of $\Omega_v$ within each cut-off. One scalar is applied to the complete prediction, so relative structure and depth discontinuities are unchanged and the metric measures the accuracy of the rendered geometry. Cut-offs at 10, 20, and 30\,m show how this accuracy develops as progressively more distant structure, which produces smaller image displacements and fewer events, enters the evaluation. Because the valid set differs across cut-offs, values are compared within each distance column.

        \begin{table*}[t]
        \centering
        \scriptsize
        \setlength{\tabcolsep}{3.5pt}
        \renewcommand{\arraystretch}{1.2}
        \caption{Mean absolute depth error [m] at 10/20/30\,m cut-offs. Best in \textbf{bold}, second-best underlined, -- not reported. EMoDepth (E) is trained with event voxels and (E+I) with intensity frames~\cite{zhu2024unsup}. The shaded (E+I) column is a cross-modal reference and is not ranked.}
            \resizebox{\textwidth}{!}{%
                \begin{tabular}{l >{\columncolor{black!6}}c >{\columncolor{black!6}}c >{\columncolor{black!6}}c c c c c c c c c c ccc}
                    \toprule
                    \multirow{1}{*}{Seq.} &
                    \multicolumn{3}{>{\columncolor{black!6}}c}{EMoDepth (E+I) \cite{zhu2024unsup}} &
                    \multicolumn{3}{c}{EMoDepth (E) \cite{zhu2024unsup}} &
                    \multicolumn{3}{c}{\citet{zhu2019unsup}} &
                    \multicolumn{3}{c}{\citet{hagenaars2025ondeviceselfsup}} &
                    \multicolumn{3}{c}{\textbf{Ours}} \\
                    \cmidrule(lr){2-4}
                    \cmidrule(lr){5-7}
                    \cmidrule(lr){8-10}
                    \cmidrule(lr){11-13}
                    \cmidrule(lr){14-16}
                    & 10m & 20m & 30m & 10m & 20m & 30m & 10m & 20m & 30m & 10m & 20m & 30m & 10m & 20m & 30m \\
                    \midrule
                    outdoor\_day1
                    & 1.40 & 2.07 & 2.65 & 3.90 & 3.79 & 4.89 & 2.72 & 3.84 & 4.40 & \underline{2.25} & \underline{3.36} & \underline{4.23} & \textbf{2.20} & \textbf{2.81} & \textbf{4.00} \\
                    \midrule
                    outdoor\_night1
                    & 2.18 & 2.70 & 3.64 & 5.55 & 4.57 & 5.72 & \underline{3.13} & 4.02 & 4.89 & 3.25 & \underline{3.83} & \underline{4.50} & \textbf{2.28} & \textbf{2.89} & \textbf{4.15} \\
                    \midrule
                    outdoor\_night2
                    & 2.06 & 2.76 & 3.42 & 5.76 & 4.48 & 5.26 & \textbf{2.19} & \underline{3.15} & \textbf{3.92} & -- & -- & -- & \underline{2.40} & \textbf{2.96} & \underline{4.02} \\
                    \midrule
                    outdoor\_night3
                    & 2.09 & 2.82 & 3.52 & 5.87 & \underline{4.39} & 5.33 & \underline{2.86} & 4.46 & \underline{5.05} & -- & -- & -- & \textbf{2.59} & \textbf{3.08} & \textbf{4.20} \\
                    \bottomrule
                \end{tabular}
            }
            \label{tab:mae_cutoffs_scaled_median}
        \end{table*}

    \subsection{Depth Results}
        Table~\ref{tab:mae_cutoffs_scaled_median} tests the first two claims together, because each scored map is produced by self-supervised per-event depth followed by occlusion-aware rendering. Our method is the most accurate event-only method at every 20\,m cut-off and on three of the four 10\,m and 30\,m cut-offs. Relative to the strongest event-only alternative at each cut-off, its error is 6.0--29.8\% lower at 20\,m. At 10\,m and 30\,m it is up to 27.2\% and 16.8\% lower, with \citet{zhu2019unsup} ahead on \texttt{outdoor\_night2} at both. The largest margin occurs on \texttt{outdoor\_night3}, where the 20\,m error is 29.8\% below the next-best event-only result.

        The lead at the wider cut-offs is consistent with the continuous-time formulation. Distant structure produces smaller image displacements and fewer events, so its evidence is easily diluted when events are accumulated over a window. Predicting depth for each event at its own timestamp and resolving collisions only at render time retains these weak motion cues. The night sequences show the value of the event-derived training signal most clearly. Relative to EMoDepth (E), the error of our method is approximately 21--59\% lower across the night cut-offs, and its night errors remain close to its daytime errors, consistent with a training signal derived from event alignment rather than from image formation under low light. EMoDepth (E+I), which also infers depth from events alone but uses intensity frames in its training loss, records lower errors. On the three night sequences, our method comes within 0.10--0.68\,m of it at every cut-off.

        \providecommand{\QualRangeOption}{own}
        \begin{figure*}[t]
            \centering
            \begingroup
            \newcommand{\PanelImg}[1]{{\setlength{\fboxsep}{0pt}\setlength{\fboxrule}{0.5pt}\fcolorbox{black!40}{white}{\includegraphics[width=0.218\textwidth]{#1}}}}
            \setlength{\tabcolsep}{3.5pt}
            \renewcommand{\arraystretch}{0.92}
            \begin{tabular}{@{}ccccc@{}}
                \makebox[0.045\textwidth]{} & \makebox[0.22\textwidth][c]{\bfseries\sffamily\small Reference image} & \makebox[0.22\textwidth][c]{\bfseries\sffamily\small Event image} & \makebox[0.22\textwidth][c]{\bfseries\sffamily\small Ground-truth depth} & \makebox[0.22\textwidth][c]{\bfseries\sffamily\small Predicted depth} \\

                \raisebox{-.5\height}{\makebox[0.045\textwidth][c]{\rotatebox[origin=c]{90}{\shortstack{\bfseries\sffamily Outdoor day\\[-1pt]\normalfont\scriptsize outdoor\_day1}}}} & \raisebox{-.5\height}{\PanelImg{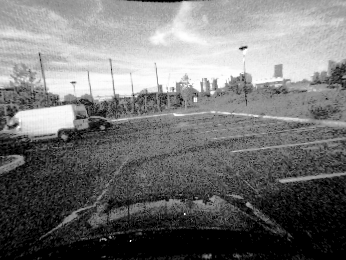}} & \raisebox{-.5\height}{\PanelImg{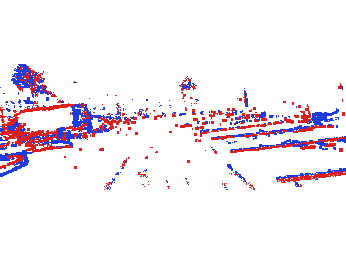}} & \raisebox{-.5\height}{\PanelImg{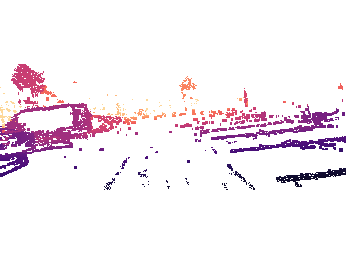}} & \raisebox{-.5\height}{\PanelImg{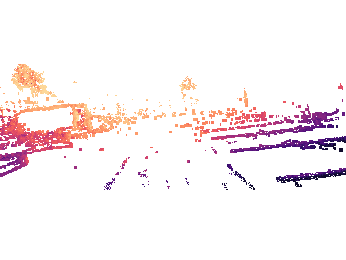}} \\[5pt]

                \raisebox{-.5\height}{\makebox[0.045\textwidth][c]{\rotatebox[origin=c]{90}{\shortstack{\bfseries\sffamily Outdoor night\\[-1pt]\normalfont\scriptsize outdoor\_night1}}}} & \raisebox{-.5\height}{\PanelImg{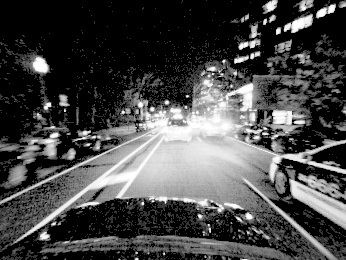}} & \raisebox{-.5\height}{\PanelImg{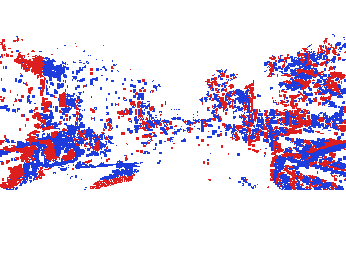}} & \raisebox{-.5\height}{\PanelImg{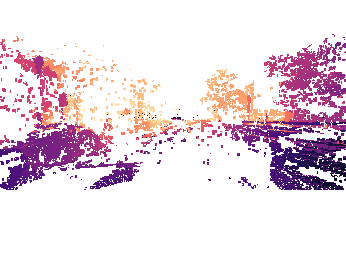}} & \raisebox{-.5\height}{\PanelImg{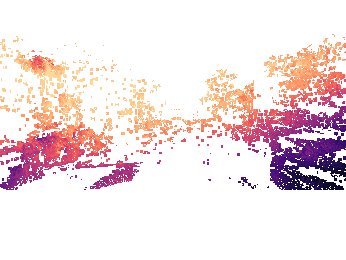}} \\
            \end{tabular}
            \endgroup
            \caption{Qualitative results on \texttt{outdoor\_day1} (top) and \texttt{outdoor\_night1} (bottom): reference image (not used by the model), input events, ground-truth depth where our depth exists, and depth at the reference time.}
            \label{fig:best_predictions_grid_panels}
        \end{figure*}

        \paragraph{Qualitative behaviour.} Figure~\ref{fig:best_predictions_grid_panels} complements the scalar metric. Depth exists only where transported events land, so it follows the scene's edges: on \texttt{outdoor\_day1} the parked van, the lamp posts and the lane markings are ordered in depth as in the ground truth, while the textureless road surface produces few events. On \texttt{outdoor\_night1} the conventional image is degraded by low light and glare, yet the event stream still carries the edges of the parked cars, the lane markings, and the street-side buildings, and the predicted depth recovers their layout, consistent with the low night-time errors in Table~\ref{tab:mae_cutoffs_scaled_median}. In both, however, the far end of the street is placed too close: the median predicted depth stays within 1\,m of the ground truth up to 20\,m, but beyond 20\,m it saturates at 11--16\,m, even where the true depth is 34--37\,m.

    \subsection{Ego-Motion}
    \label{sec:experiments:egomotion}
        \begin{table}[t]
            \centering
            \scriptsize
            \setlength{\tabcolsep}{3pt}
            \caption{Ego-motion on MVSEC, ARPE [deg] / ARRE [$10^{-3}$\,rad]. Lower is better, best in \textbf{bold}, -- not reported. Ours: motion over each 50\,ms depth interval.}
            \label{tab:egomotion}
            \begin{tabular}{@{}lcccc@{}}
                \toprule
                Method & day1 & night1 & night2 & night3 \\
                \midrule
                \citet{zhu2019unsup} & \textbf{7.74} / 8.67 & -- & -- & -- \\
                \textbf{Ours} & 7.80 / \textbf{5.67} & 3.48 / 3.64 & 5.57 / 4.46 & 5.19 / 4.85 \\
                \bottomrule
            \end{tabular}
        \end{table}

        At test time, the ego-motion estimate for each window is the constant 6-DoF velocity that best aligns its events under the predicted per-event depths. It is also the training motion of the depth losses and the camera-motion head's target. Its accuracy therefore depends on the predicted 3D event locations. The estimate is then refined in two steps. Temporal smoothing averages each estimate with up to four preceding ones. A vehicle prior uses the non-holonomic motion of a car: the translation follows the vehicle's longitudinal axis, which is fixed in the frame of a rigidly mounted camera and is estimated once on the training sequence. Each translation keeps its estimated speed and sign. Among the depth baselines, only \citet{zhu2019unsup} report ego-motion, on \texttt{outdoor\_day1} alone, because their network failed under the flashing street lights of the night sequences. Table~\ref{tab:egomotion} uses their metrics: ARPE, the angle between the predicted and ground-truth translation directions, and ARRE, the rotation-error angle.

        On \texttt{outdoor\_day1}, the rotation error of our method is 35\% below that of \citet{zhu2019unsup} ($5.67$ vs $8.67\times10^{-3}$\,rad), and its translation-direction error is similar to theirs (7.80 vs 7.74\,deg). Rotation is the more informative of the two comparisons: on this driving data the translation direction is dominated by forward motion, and in our method it follows the vehicle prior's axis, whereas the rotation must be recovered from how the transported events align. The lower rotation error therefore extends the depth result of Table~\ref{tab:mae_cutoffs_scaled_median} to the camera motion: the per-event depths that render accurate target-time maps also place the events consistently enough in 3D for their alignment to determine the camera's rotation. The errors do not increase in low light: on the three night sequences, for which \citet{zhu2019unsup} report no result, ARRE is $3.64$--$4.85\times10^{-3}$\,rad and ARPE 3.48--5.57\,deg, both below the daytime values.

    \subsection{Ablations}
        \paragraph{Local spatiotemporal encoder.} The third claim is examined first by removing the local branch, so that only the selective state-space path remains. This tests the premise of the local-global design, that the ordered event sequence alone does not represent image-plane neighbourhoods. Without the local branch the depth error on \texttt{outdoor\_day1} rises substantially, so long-range temporal context does not substitute for local geometric context: depth from motion parallax depends on how an edge and its neighbourhood move together on the image plane, and on whether nearby events belong to the same surface or lie on either side of an occlusion boundary. These relations are expressed by spatial adjacency rather than by sequence order, and the local branch supplies them at several scales to each event.

        \paragraph{Temporal backbone.} The selective state-space backbone is then compared with GRU, LSTM, and vanilla RNN alternatives on the same event windows, each run in one direction at a matched parameter budget. The state-space backbone processes the windows faster than every alternative, which supports selective state-space modelling for long event sequences. Together, the backbone comparison and local-branch ablation support combining efficient long-range temporal context with local spatial structure.

        \paragraph{Supervision.} Finally, the self-supervised objective is examined with events alone. The temporal-dispersion and out-of-bounds terms are removed in turn and together, which tests whether event sharpness by itself constrains the per-event depths, and event sharpness is replaced with soft contrast maximisation \citep{gallego2018contrast,hagenaars2025ondeviceselfsup}, which tests what the dual-anchor terms add to an established event-alignment objective. Per-sequence depth and ego-motion errors of all ablations are given in the supplementary material.

\section{Conclusion}
    We presented a self-supervised continuous-time formulation for event-native depth estimation that also recovers ego-motion from the same events. Per-event depth is predicted by a local-global event-state encoder, learned from dual-anchor motion-compensated event alignment, and rendered to target times by a differentiable occlusion-aware renderer. Geometric transport provides the supervisory signal by rewarding depth predictions that explain the observed event motion under the estimated camera motion. On MVSEC, our method is the most accurate event-only self-supervised method at every 20\,m cut-off and at three of the four 10\,m and 30\,m cut-offs, with its largest margins at night, and the ablations attribute its accuracy to both the local branch and the alignment objective. Camera motion recovered from the same events using these depths has a rotation error 35\% below that of \citet{zhu2019unsup} on \texttt{outdoor\_day1}.

\FloatBarrier

{
    \small
    \bibliographystyle{ieeenat_fullname}
    \bibliography{main}
}

\clearpage
\setcounter{page}{1}
\maketitlesupplementary
\appendix          %
\raggedbottom

\section{Method (Section~3)}
\label{sec:supp:method}

\paragraph{Encoder and depth head (Section~3.2, Figure~\ref{fig:blockflow}).} Table~\ref{tab:supp:impl} lists the settings of the model in the notation of Section~3.2: the width $F$ of the tokens $\mathbf{h}_i^0$, the number of temporal bins and the spatial stride of the voxelisation $\mathcal{V}^{(r)}$ at each of the three local scales, and the bounds of the depth head of Equation~\eqref{eq:depth-head}, which cover the range scored in Table~\ref{tab:mae_cutoffs_scaled_median}. The network is initialised randomly. %

\paragraph{Camera-motion head and training motion (Sections~3.2 and~\ref{sec:losses}).} The best-aligning velocity $\bar{\boldsymbol{\xi}}$, which transports the events in Equation~\eqref{eq:total-loss} and is the target of the camera-motion head, minimises $\mathcal{L}_{\mathrm{CMax}}$ of Equation~\eqref{eq:supp:cmax} under the current depths, from a coarse grid of forward speeds and yaw rates the first time a window is seen and from its previous solution afterwards (Table~\ref{tab:supp:impl}). %
The gauge term of Section~\ref{sec:losses} is $\mathcal{L}_{\mathrm{gauge}}=(\overline{\log\hat d}-\log 10)^2$, where $\overline{\log\hat d}$ is the batch mean of each window's mean log depth (depth in metres); added to Equation~\eqref{eq:total-loss} with weight 1, it fixes only the global scale. %
The checkpoint with the lowest held-out value of Equation~\eqref{eq:total-loss}, without the gauge, on 64 windows of \texttt{outdoor\_day2} is kept. %

\paragraph{Soft contrast maximisation (Sections~\ref{sec:losses} and~4.4).} The velocity $\bar{\boldsymbol{\xi}}$ minimises a soft multi-reference contrast-maximisation objective~\cite{gallego2018contrast,hagenaars2025ondeviceselfsup}, also used in place of $\mathcal{L}_{\mathrm{sharp}}$ in Section~\ref{sec:supp:ablations}. Instead of the anchors $\mathcal{A}$ (Section~\ref{sec:losses}), events are warped with Equation~\eqref{eq:warp} to 13 equally spaced times $\mathcal{A}_{13}$ from $t_0$ to $t_1$, and events that leave the image or pass behind the camera at any of them are dropped. With $b^{a}_{\mathbf{u},i}$ as in Equation~\eqref{eq:iwe}, $\tau_a=(a-t_0)/(t_1-t_0)$ and $\eta_{a,i}=|\tau_i-\tau_a|$ ($\tau_i$ of Section~\ref{sec:problem}), each polarity $p\in\{-1,+1\}$ gives a count image and an average-time image,
\begin{equation}
    C_{a,p}(\mathbf{u})=\sum_{i:\,p_i=p} b^{a}_{\mathbf{u},i},\qquad
    \bar{\eta}_{a,p}(\mathbf{u})=\frac{1}{C_{a,p}(\mathbf{u})}\sum_{i:\,p_i=p} b^{a}_{\mathbf{u},i}\,\eta_{a,i},
\end{equation}
and the objective is
\begin{equation}
    \mathcal{L}_{\mathrm{CMax}}=\frac{1}{13}\sum_{a\in\mathcal{A}_{13}}\frac{\sum_{p,\mathbf{u}} \nu_{a,p}(\mathbf{u})\,\bar{\eta}_{a,p}(\mathbf{u})^{2}}{\sum_{p,\mathbf{u}} \nu_{a,p}(\mathbf{u})},
    \nu_{a,p}=\frac{C_{a,p}}{C_{a,p}+1}.
    \label{eq:supp:cmax}
\end{equation}
It is low when events of different times land on the same pixels, that is, when the warped events are sharp. The weight $\nu_{a,p}$ lets pixels reached by a sliver of an event contribute little, so $\mathcal{L}_{\mathrm{CMax}}$ is continuous in the warped positions. %

\paragraph{Objective weights (Equation~\eqref{eq:total-loss}).} The model uses both anchors $t_0$ and $t_1$ and the weights of Table~\ref{tab:supp:impl}, without the smoothness term $\mathcal{L}_{\mathrm{smooth}}$ of Equation~\eqref{eq:smooth}, so that Tables~\ref{tab:supp:egomotion} and~\ref{tab:supp:ablations} isolate event alignment. %

\begin{table}[t]
    \centering
    \scriptsize
    \setlength{\tabcolsep}{4pt}
    \caption{Settings of the model.}
    \label{tab:supp:impl}
    \begin{tabular}{@{}L{0.45\linewidth}L{0.48\linewidth}@{}}
        \toprule
        Component & Setting \\
        \midrule
        Token width $F$ / temporal stage & 64 / bidirectional Mamba~\cite{gu2024mamba} \\ %
        $\mathcal{V}^{(r)}$: temporal bins / stride & 32 / 3, 64 / 5, 128 / 9\,px \\ %
        Depth-network parameters & 3.59\,M \\ %
        Window / events per 50\,ms interval & two 50\,ms intervals / 7,500 \\ %
        Depth bounds $[d_{\min},d_{\max}]$ & $[0.25,30]$\,m \\ %
        $\lambda_{\mathrm{sharp}},\lambda_{\mathrm{time}},\lambda_{\mathrm{oob}}$ / $\lambda_{\mathrm{smooth}}$ & 1.0, 0.5, 0.1 / 0 \\ %
        Training / held-out windows & 1,687 / 64 \\ %
        Optimiser, learning rate (head), decay & AdamW, $10^{-4}$ ($3\times10^{-4}$), $10^{-4}$, one-cycle \\ %
        Windows per step, epochs (steps) & 2, 20 (16,880) \\ %
        Solver steps (first / later visit) & 40 / 15 \\ %
        $\mathcal{L}_{\mathrm{gauge}}$: weight / target & 1 / $\log 10$ \\ %
        \bottomrule
    \end{tabular}
\end{table}

\section{Experimental Setup (Section~4.1)}
\label{sec:supp:setup}

\begin{figure}[t]
    \centering
    \includegraphics[width=0.92\linewidth]{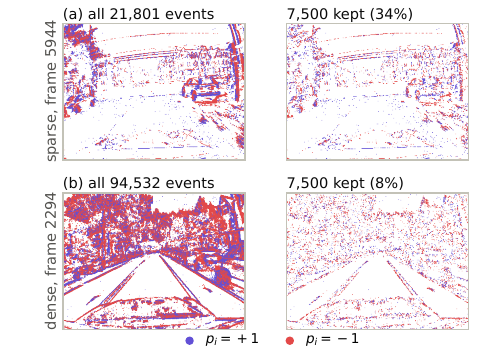}
    \caption{Event subsampling on two training intervals of \texttt{outdoor\_day2}, a sparse one (a; 21{,}801 events, the 5th percentile of the training intervals) and a dense one (b; 94{,}532 events, the 95th percentile). Each 50\,ms interval is reduced on its own to 7{,}500 events, drawn evenly over 30 time bins and a $16\times12$ grid. Left: all events; right: kept events; colour: polarity.}
    \label{fig:supp:subsampling}
\end{figure}

\paragraph{Dataset and windows.} A window of the model spans two consecutive 50\,ms intervals, and each interval is reduced on its own to 7,500 events, so that a window has the same cost whatever the scene texture or illumination. %
The time span of an interval is divided into 30 equal bins and the image into a $16\times12$ grid. Every occupied time--image cell receives the same quota of uniformly drawn events (all of them when it holds fewer), and the remaining budget goes to randomly chosen cells that still hold events. %
This is the time-stratified subsampling of Section~4.1; the selected events keep their exact pixel coordinates, timestamps and polarities, which the continuous-time warp of Equation~\eqref{eq:warp} requires. Figure~\ref{fig:supp:subsampling} shows a sparse and a dense training interval (5th and 95th percentiles of the event count); each keeps 7,500 events, 34\% and 8\% of the interval, and every occupied time--image cell keeps at least one. In the dense interval the events per time bin range from 1,351 to 4,659 but the kept events only from 208 to 278, so the alignment terms of Section~\ref{sec:losses} are evaluated along the whole interval rather than at its densest instants. %

\paragraph{Metric and scale alignment (Table~\ref{tab:mae_cutoffs_scaled_median}).} Table~\ref{tab:mae_cutoffs_scaled_median} scores every ground-truth frame. %
The map scored at a frame is rendered from the events of the 50\,ms interval that ends at the frame, transported with the camera-motion head's motion. Image rows from 190 downwards, which show the vehicle hood, are excluded, and the set $\Omega_v$ of Equation~\eqref{eq:median-scale} then holds 2,570, 2,686, 3,342 and 3,263 pixels per frame on average. Each frame has its own factor $s$ of Equation~\eqref{eq:median-scale}, and the absolute errors up to 10, 20 and 30\,m are pooled over all pixels of all frames. %

\section{Ego-Motion (Section~\ref{sec:experiments:egomotion})}
\label{sec:supp:egomotion}

\paragraph{Estimates of Table~\ref{tab:supp:egomotion}.} Let $\boldsymbol{\xi}_j=(\boldsymbol{\omega}_j,\mathbf{v}_j)$ be the angular and linear velocity, in the camera frame, of the window that ends at ground-truth frame $t_j$. Its relative pose $(\hat{\mathbf{R}},\hat{\mathbf{t}})$ over the last 50\,ms interval is compared with the ground truth $(\mathbf{R},\mathbf{t})$~\cite{zhu2019unsup}: ARPE is the mean angle between $\hat{\mathbf{t}}$ and $\mathbf{t}$, and ARRE the mean rotation angle of $\hat{\mathbf{R}}^{\top}\mathbf{R}$; frames whose neighbouring ground-truth poses are more than 0.1\,s apart are not scored. The rows of Table~\ref{tab:supp:egomotion} are four estimates of $\boldsymbol{\xi}_j$. The camera-motion head and the alignment solver are separate estimates; a leading $+$ applies a step to the row above, so `$+$ vehicle axis prior' is the smoothed solver estimate with the prior. %

\emph{Camera-motion head:} the velocity regressed in one forward pass by the head of Section~3.2. It is trained towards $\bar{\boldsymbol{\xi}}$ (Section~\ref{sec:supp:method}), transports the events scored in Table~\ref{tab:mae_cutoffs_scaled_median} and is the ego-motion scored in Table~\ref{tab:supp:ablations}. %

\emph{Alignment solver:} the estimate of Section~\ref{sec:experiments:egomotion}, the best-aligning velocity $\bar{\boldsymbol{\xi}}_j$ of Section~\ref{sec:losses}, computed at test time from the window's events and predicted depths alone. It minimises $\mathcal{L}_{\mathrm{CMax}}$ of Equation~\eqref{eq:supp:cmax} with count weights $\nu_{a,p}=C_{a,p}$ by 60 gradient steps from the best of a coarse grid of forward speeds and yaw rates. It is iterative where the head needs one pass, but it uses the predicted depths directly. %

\begin{table}[t]
    \centering
    \scriptsize
    \setlength{\tabcolsep}{3pt}
    \caption{Ego-motion of the model after each step of Section~\ref{sec:experiments:egomotion} (a leading $+$ adds a step to the row above), with the values of Table~\ref{tab:egomotion}. ARPE [deg] / ARRE [$10^{-3}$\,rad] over the last 50\,ms interval, all frames. Best in \textbf{bold}.}
    \label{tab:supp:egomotion}
    \adjustbox{max width=\linewidth}{\begin{tabular}{@{}lcccc@{}}
\toprule
Method & Day 1 & Night 1 & Night 2 & Night 3 \\
\midrule
\citet{zhu2019unsup} (Table~\ref{tab:egomotion}) & 7.74 / 8.67 & -- & -- & -- \\
Ours (Table~\ref{tab:egomotion}) & 7.80 / 5.67 & 3.48 / 3.64 & 5.57 / 4.46 & 5.19 / 4.85 \\
\midrule
Camera-motion head & 7.66 / 6.68 & 4.13 / 4.35 & 5.94 / 6.19 & 5.58 / 6.31 \\
Alignment solver & 12.42 / 5.66 & 9.51 / 3.59 & 7.70 / 4.33 & 7.73 / \textbf{4.68} \\
$+$ temporal smoothing & 11.35 / \textbf{5.51} & 9.11 / \textbf{3.47} & 7.37 / \textbf{4.27} & 7.38 / 4.71 \\
$+$ vehicle axis prior & \textbf{6.49} / \textbf{5.51} & \textbf{2.70} / \textbf{3.47} & \textbf{4.87} / \textbf{4.27} & \textbf{4.44} / 4.71 \\
\bottomrule
\end{tabular}
}
\end{table}

\begin{table*}[t]
    \centering
    \scriptsize
    \setlength{\tabcolsep}{4pt}
    \caption{Ablations of the model (Section~\ref{sec:supp:ablations}). Depth: MAE [m] under the protocol of Table~\ref{tab:mae_cutoffs_scaled_median} on \texttt{outdoor\_day1} and the mean of the three nights. Ego-motion: ARPE [deg] and ARRE [$10^{-3}$\,rad] of the camera-motion head, mean of the four sequences. Parameter counts in parentheses. Each group starts with the full model; lower is better, best per column within each group in \textbf{bold}.}
    \label{tab:supp:ablations}
    \begin{tabular}{@{}lccc@{\hspace{10pt}}ccc@{\hspace{10pt}}cc@{}}
\toprule
 & \multicolumn{3}{c}{\texttt{outdoor\_day1} MAE [m]} & \multicolumn{3}{c}{Nights MAE [m]} & \multicolumn{2}{c}{Camera-motion head} \\
\cmidrule(lr){2-4}\cmidrule(lr){5-7}\cmidrule(l){8-9}
Variant & 10\,m & 20\,m & 30\,m & 10\,m & 20\,m & 30\,m & ARPE & ARRE \\
\midrule
\multicolumn{9}{@{}l}{\textit{Local spatiotemporal encoder}} \\
Full model (ours, 3.59\,M) & \textbf{2.24} & \textbf{2.89} & \textbf{4.06} & \textbf{2.40} & \textbf{3.10} & \textbf{4.31} & 5.83 & \textbf{5.88} \\
Without the local branch (0.45\,M) & 2.38 & 3.12 & 4.14 & 2.65 & 3.30 & \textbf{4.31} & \textbf{5.50} & 5.96 \\
\midrule
\multicolumn{9}{@{}l}{\textit{Temporal backbone}} \\
Full model (ours, bidirectional Mamba, 3.59\,M) & 2.24 & 2.89 & 4.06 & 2.40 & 3.10 & 4.31 & 5.83 & 5.88 \\
Mamba, one direction (3.38\,M) & 2.20 & 2.89 & 4.11 & 2.39 & 3.13 & 4.41 & \textbf{5.01} & \textbf{5.68} \\
GRU, one direction (3.33\,M) & 2.26 & 2.95 & 4.14 & 2.38 & 3.13 & 4.41 & 5.49 & 5.73 \\
LSTM, one direction (3.38\,M) & 2.22 & 2.91 & 4.13 & 2.38 & 3.11 & 4.37 & 5.48 & 5.99 \\
Vanilla RNN, one direction (3.23\,M) & \textbf{2.14} & \textbf{2.79} & \textbf{4.01} & \textbf{2.37} & \textbf{3.07} & \textbf{4.29} & 6.10 & 6.04 \\
\midrule
\multicolumn{9}{@{}l}{\textit{Supervision}} \\
Full model (ours) & \textbf{2.24} & \textbf{2.89} & \textbf{4.06} & \textbf{2.40} & \textbf{3.10} & \textbf{4.31} & 5.83 & 5.88 \\
Without $\mathcal{L}_{\mathrm{time}}$ & 2.26 & 2.97 & 4.16 & 2.42 & 3.14 & 4.40 & 5.51 & 5.75 \\
Without $\mathcal{L}_{\mathrm{oob}}$ & 2.43 & 3.08 & 4.08 & 2.68 & 3.33 & 4.32 & \textbf{5.01} & \textbf{5.53} \\
$\mathcal{L}_{\mathrm{sharp}}$ only & 2.32 & 2.97 & \textbf{4.06} & 2.54 & 3.23 & 4.35 & 5.88 & 5.99 \\
Soft CMax in place of $\mathcal{L}_{\mathrm{sharp}}$ & 4.38 & 4.52 & 5.48 & 4.60 & 4.80 & 5.76 & 5.60 & 6.11 \\
Soft CMax, without $\mathcal{L}_{\mathrm{oob}}$ & 2.51 & 3.25 & 4.41 & 2.63 & 3.46 & 4.69 & 6.14 & 5.70 \\
Soft CMax, without $\mathcal{L}_{\mathrm{time}}$ & 4.74 & 4.73 & 5.79 & 4.73 & 4.80 & 5.93 & 7.80 & 6.07 \\
Warp to $t_1$ only & 2.76 & 3.41 & 4.54 & 3.36 & 3.90 & 4.94 & 6.22 & 5.66 \\
\midrule
\multicolumn{9}{@{}l}{\textit{Training motion}} \\
Full model (ours) & 2.24 & 2.89 & \textbf{4.06} & \textbf{2.40} & \textbf{3.10} & \textbf{4.31} & 5.83 & \textbf{5.88} \\
Camera-motion head's velocity & \textbf{2.19} & \textbf{2.88} & 4.08 & \textbf{2.40} & 3.16 & 4.40 & \textbf{5.66} & 5.96 \\
\midrule
\multicolumn{9}{@{}l}{\textit{Window length}} \\
Full model (ours) & \textbf{2.24} & \textbf{2.89} & \textbf{4.06} & \textbf{2.40} & \textbf{3.10} & \textbf{4.31} & 5.83 & 5.88 \\
Three 50\,ms intervals per window & 2.29 & 3.03 & 4.15 & 2.56 & 3.31 & 4.47 & \textbf{5.46} & \textbf{5.73} \\
\bottomrule
\end{tabular}

\end{table*}

\emph{$+$ temporal smoothing:} a car's velocity changes little within a fraction of a second, so $\bar{\boldsymbol{\xi}}_j$ is replaced by the mean $\tilde{\boldsymbol{\xi}}_j$ of the $\bar{\boldsymbol{\xi}}_k$ with $t_j-0.225\,\mathrm{s}\le t_k\le t_j$, the current window and up to four preceding ones at 20\,Hz. Only past windows are used, so the step runs online. %

\emph{$+$ vehicle axis prior:} within a window, translation moves the events little and by an amount that scales with the inverse depth, whereas rotation moves every event regardless of its depth, so the translation direction is the less constrained part of the estimate. For a camera rigidly mounted on a car, it follows the vehicle's longitudinal axis $\mathbf{n}$, which is fixed in the camera frame (Section~\ref{sec:experiments:egomotion}). The axis is estimated once, without labels, as the normalised mean of $\mathbf{v}_k/\|\mathbf{v}_k\|$ over the camera-motion head's velocities on \texttt{outdoor\_day2} with above-median speed $\|\mathbf{v}_k\|$ (no near-standstill windows). Each $\tilde{\mathbf{v}}_j$ becomes $\operatorname{sign}(\mathbf{n}^{\top}\tilde{\mathbf{v}}_j)\,\|\tilde{\mathbf{v}}_j\|\,\mathbf{n}$, keeping its speed and sign; the rotation is unchanged. %

\paragraph{Model (Table~\ref{tab:supp:egomotion}).} The alignment solver lowers the rotation error of the camera-motion head by 15--30\%, while its translation direction is less accurate (7.70--12.42$^\circ$ against 4.13--7.66$^\circ$). Temporal smoothing lowers the translation-direction error on every sequence and the rotation error on three, and the vehicle prior brings the translation-direction error to 2.70--6.49$^\circ$ without changing the rotation. The errors are then below those of Table~\ref{tab:egomotion} on every sequence and, on \texttt{outdoor\_day1}, below those of \citet{zhu2019unsup}. %

\section{Ablations (Section~4.4)}
\label{sec:supp:ablations}

\paragraph{Protocol.} Table~\ref{tab:supp:ablations} reports the ablations of Section~4.4 and further design choices. Each row changes one component of the model and is trained from scratch with the same data, schedule, seed and checkpoint rule. %
Depth is scored exactly as in Table~\ref{tab:mae_cutoffs_scaled_median}, so the full model's 2.24, 2.89 and 4.06\,m on \texttt{outdoor\_day1} compare directly with the 2.20, 2.81 and 4.00\,m of the paper's model; ego-motion is that of the camera-motion head before the refinement steps of Table~\ref{tab:supp:egomotion}, with ARPE within 5.0--6.2$^\circ$ in all rows but one discussed below and ARRE within 5.5--$6.1\times10^{-3}$\,rad. %

\paragraph{Local spatiotemporal encoder.} Removing the local branch, which leaves the selective state-space path alone (Section~4.4), raises the \texttt{outdoor\_day1} error from 2.24, 2.89 and 4.06\,m to 2.38, 3.12 and 4.14\,m, and the night error at 10 and 20\,m, from 2.40 and 3.10\,m to 2.65 and 3.30\,m. %

\paragraph{Temporal backbone.} The backbone rows of Table~\ref{tab:supp:ablations} run Mamba, GRU, LSTM or a vanilla RNN in one direction (forward pass only) at similar parameter counts (Table~\ref{tab:supp:ablations}). %
Their depth errors lie within 0.11\,m of the full model at every cut-off, on \texttt{outdoor\_day1} and at night, so the backbones differ mainly in speed; the vanilla RNN has the highest translation-direction error of them (6.10$^\circ$). %
The throughput of Section~4.4 was measured on an RTX 2080 Ti at 30,000 events (median of 50 forward passes): ours runs at 3.30 windows per second, 18--27\% more than the recurrent ones. %

\paragraph{Supervision.} The supervision rows remove the temporal-dispersion term $\mathcal{L}_{\mathrm{time}}$ (Equation~\eqref{eq:time}) and the out-of-bounds term $\mathcal{L}_{\mathrm{oob}}$ (Equation~\eqref{eq:oob}) in turn and together (Section~4.4). Each removal raises the error at 10 and 20\,m, to 2.26--2.43 and 2.97--3.08\,m on \texttt{outdoor\_day1} and to 2.42--2.68 and 3.14--3.33\,m at night, so event sharpness $\mathcal{L}_{\mathrm{sharp}}$ (Equation~\eqref{eq:sharp}) alone constrains the per-event depths less well than the full objective. %
The soft contrast-maximisation row replaces $\mathcal{L}_{\mathrm{sharp}}$ with $\mathcal{L}_{\mathrm{CMax}}$ of Equation~\eqref{eq:supp:cmax}, computed on the predicted depths, and keeps $\mathcal{L}_{\mathrm{time}}$ and $\mathcal{L}_{\mathrm{oob}}$ at our weights; it raises the \texttt{outdoor\_day1} error to 4.38, 4.52 and 5.48\,m. Keeping only $\mathcal{L}_{\mathrm{time}}$ or only $\mathcal{L}_{\mathrm{oob}}$ beside it also stays above the full model, and without $\mathcal{L}_{\mathrm{time}}$ ARPE rises to 7.80$^\circ$. %
The dual anchors $\mathcal{A}=\{t_0,t_1\}$ of Section~\ref{sec:losses} are tested by warping to $t_1$ alone with $\mathcal{L}_{\mathrm{sharp}}$ doubled, which gives 2.76, 3.41 and 4.54\,m, so the second anchor helps. %

\paragraph{Further factors.} The remaining rows of Table~\ref{tab:supp:ablations} test choices of Sections~\ref{sec:problem}--\ref{sec:losses} that Section~4.4 does not ablate. Training the depth under the camera-motion head's velocity instead of $\bar{\boldsymbol{\xi}}$ gives 2.19, 2.88 and 4.08\,m on \texttt{outdoor\_day1}, and a window of three 50\,ms intervals gives 2.29, 3.03 and 4.15\,m, above the two-interval window at every cut-off. %

\end{document}